\documentclass[5p]{elsarticle}

\usepackage{microtype}
\usepackage{amsmath}
\usepackage{amsfonts}
\usepackage{amssymb}
\usepackage{textcomp}
\usepackage{yhmath}
\usepackage{caption}
\usepackage[usenames,dvipsnames]{color}
\usepackage{booktabs}
\usepackage{natbib}
\usepackage[colorlinks=true]{hyperref}
\usepackage{mathtools}
\usepackage{subfig}
\DeclareMathOperator*{\argmax}{arg\,max}

\DeclarePairedDelimiter{\abs}{\vert}{\vert}

\journal{Engineering Applications of Artificial Intelligence}

\begin{document}
\emergencystretch 3em
	
\begin{frontmatter}

\title{Interpretable Policies for Reinforcement Learning by Genetic Programming}
\author[tu,sie]{Daniel~Hein\corref{cor}}
\ead{daniel.hein@in.tum.de}
\author[sie]{Steffen~Udluft}
\author[tu,sie]{Thomas~A.~Runkler}

\address[tu]{Technical University of Munich, Department of Informatics, Boltzmannstr. 3, 85748 Garching, Germany}
\address[sie]{Siemens AG, Corporate Technology, Otto-Hahn-Ring 6, 81739 Munich, Germany}

\cortext[cor]{Corresponding author}

\begin{abstract}
	The search for interpretable reinforcement learning policies is of high academic and industrial interest.
	Especially for industrial systems, domain experts are more likely to deploy autonomously learned controllers if they are understandable and convenient to evaluate.
	Basic algebraic equations are supposed to meet these requirements, as long as they are restricted to an adequate complexity.
	Here we introduce the \textit{genetic programming for reinforcement learning} (GPRL) approach based on model-based batch reinforcement learning and genetic programming, which autonomously learns policy equations from pre-existing default state-action trajectory samples.
	GPRL is compared to a straight-forward method which utilizes genetic programming for symbolic regression, yielding policies imitating an existing well-performing, but non-interpretable policy.
	Experiments on three reinforcement learning benchmarks, i.e., mountain car, cart-pole balancing, and industrial benchmark, demonstrate the superiority of our GPRL approach compared to the symbolic regression method.
	GPRL is capable of producing well-performing interpretable reinforcement learning policies from pre-existing default trajectory data.
\end{abstract}


\end{frontmatter}

\section{Introduction}
\label{section:introduction}

This work introduces a genetic programming (GP) approach for autonomously learning interpretable reinforcement learning (RL) policies from previously recorded state transitions.
Despite the search of interpretable RL policies being of high academic and industrial interest, little has been published concerning human interpretable and understandable policies trained by data driven learning methods \citep{maes:12}.
Recent research results show that using fuzzy rules in batch RL settings can be considered an adequate solution to this task~\citep{hein:17c}.
However, in many cases the successful use of fuzzy rules requires prior knowledge about the shape of the membership functions, the number of fuzzy rules, the relevant state features, etc.
Moreover, for some problems the policy representation as a set of fuzzy rules might be generally unfavorable by some domain experts.
Our \textit{genetic programming for reinforcement learning} (GPRL) approach learns policy representations which are represented by basic algebraic equations of low complexity.

The GPRL approach is motivated by typical industrial application scenarios like wind or gas turbines.
For industrial systems, low-level control is realized by dedicated expert-designed controllers, which guarantee safety and stability.
However, we observed that high-level control is usually implemented by default control strategies, provided by best practice approaches or domain experts who are maintaining the system, based on personal experience and knowledge about the system's dynamics.
One reason for the lack of autonomously generated real-world controllers is that modeling system dependencies for high-level control by a first principle model is a complicated and often infeasible approach.
Since in many real-world applications such representations cannot be found, training high-level controllers has to be performed on data samples from the system. 
RL is capable of yielding high-level controllers based solely on available system data.

RL is concerned with learning a policy for a system that can be modeled as a Markov decision process~\citep{sutton:98}. 
This policy maps from system states to actions in the system. 
Repeatedly applying an RL policy generates a trajectory in the state-action space (Section~\ref{section:rl}).
Based on our experience, learning such RL controllers in a way that produces interpretable high-level controllers is of high interest, especially for real-world industry problems, since interpretable solutions are expected to yield higher acceptance from domain experts than black-box solutions.

In batch RL, we consider applications where online learning approaches, such as classical temporal-difference learning~\citep{sutton:88}, are prohibited for safety reasons, since these approaches require exploration of system dynamics. 
In contrast, batch RL algorithms generate a policy based on existing data and deploy this policy to the system after training. 
In this setting, either the value function or the system dynamics are trained using historic operational data comprising a set of four-tuples of the form (\textit{observation}, \textit{action}, \textit{reward}, \textit{next observation}), which is referred to as a data batch.
Research from the past two decades~\citep{gordon:95,ormoneit:02,lagoudakis:03,ernst:05} suggests that such batch RL algorithms satisfy real-world system requirements, particularly when involving neural networks (NNs) modeling either the state-action value function~\citep{riedmiller:051,riedmiller:05,schneegass:07a,schneegass:07b,riedmiller:09} or system dynamics~\citep{bakker:04,schafer:08,depeweg:16}. 
Moreover, batch RL algorithms are data-efficient~\citep{riedmiller:051,schaefer:07} because batch data is utilized repeatedly during the training phase.

To the best of our knowledge, GP-generated policies have never been combined with a model-based batch RL approach (Section~\ref{section:related}). 
In the proposed GPRL approach, the performance of a population of basic algebraic equations is evaluated by testing the individuals on a world model using the Monte Carlo method~\citep{sutton:98}.
The combined return value of a number of action sequences is the fitness value that is maximized iteratively from GP generation to generation.
	
GPRL is a novel model-based RL approach, i.e., training is conducted on an environment approximation referred to as world model.
Generating a world model from real system data in advance and training a GP policy using this model has several advantages. 
(i) In many real-world scenarios, data describing system dynamics is available in advance or is easily collected. 
(ii) Policies are not evaluated on the real system, thereby avoiding the detrimental effects of executing a bad policy. 
(iii) Expert-driven reward function engineering, yielding a closed-form differentiable equation, utilized during policy training is not required, i.e., it is sufficient to sample from the system's reward function and model the underlying dependencies by using supervised machine learning.

The remainder of this paper is organized as follows.
The RL and GP methods employed in our framework are reviewed in Sections~\ref{section:rl} and \ref{section:gp}.
Specifically, the problem of finding policies via RL is formalized as an optimization task. 
In addition, GP in general and the specific implementation that we used for experiments are motivated and presented.
An overview of how the proposed GPRL approach is derived from different methods is given in Section~\ref{section:gprl}.
Experiments using three benchmark problems, i.e., the mountain car (MC) problem, the cart-pole balancing (CPB) task, and the industrial benchmark (IB), are described in Section~\ref{section:experiments}.
Experimental results are discussed in Section~\ref{section:results}. 
The results demonstrate that the proposed GPRL approach can solve the benchmark problems and is able to produce interpretable RL policies.
To benchmark GPRL, we compare the obtained results to an alternative approach in which GP is used to mimic an existing non-interpretable NN policy by symbolic regression.

\section{Related Work}
\label{section:related}

GP has been utilized for creating rule-based policies since its introduction by \cite{koza:92}. 
Since then, the field of GP has grown significantly and has produced numerous results that can compete with human-produced results, including controllers, game playing, and robotics \citep{koza:10}. 
\cite{keane:02} automatically synthesized a controller by using GP, outperforming conventional PID controllers for an industrially representative set of plants. 
Another approach using genetic algorithms for RL policy design is to learn a set of fuzzy ``if-then" rules, by modifying membership functions, rule sets and consequent types \citep{juang:00}. 
Recently, \cite{koshiyama:14} introduced GPFIS, a genetic fuzzy controller based on multi-gene GP, and demonstrated the superiority in relation to other genetic fuzzy controllers on the cart-centering and the inverted pendulum problems. 
On the same benchmark, a movable inverted pendulum, \cite{shimooka:98} applied GP to generate equations for calculating the control force by evaluating the individuals' performances on predefined fitness functions. 

A fundamental drawback with all of the former methods is that in many real-world scenarios such dedicated expert generated fitness functions do not exist. 
In RL the goal is to derive well-performing policies only by 
(i) interacting with the environment, or by 
(ii) extracting knowledge out of pre-generated data, running the system with an arbitrary policy~\citep{sutton:98}. 
(i) is referred to as the online RL problem, for which Q-learning methods are known to produce excellent results. 
For (ii), the off\/line RL problem, model-based algorithms are usually more stable and yield better performing policies~\citep{hein:17c}.

GP in conjunction with online RL Q-learning has been used in~\citep{downing:01} on standard maze search problems and in~ \citep{kamio:05} to enable a real robot to adapt its action to a real environment. 
\cite{katagiri:02} introduced genetic network programming (GNP), which has been applied to online RL in \citep{mabu:02} and improved by Q-tables in \citep{mabu:04}. 
In these publications, the efficiency of GNP for generating RL policies has been discussed. 
This performance gain, in comparison to standard GP, comes at the cost of interpretability, since complex network graphs have to be traversed to compute the policy outputs.

\cite{gearhart:03} examined GP as a policy search technique for Markov Decision Processes. 
Given a simulation of the Freecraft tactical problem, he performed Monte Carlo simulations to evaluate the fitness of each individual. 
Note that such exact simulations are usually not available in industry.
Similarly, in \citep{maes:12} Monte Carlo simulations have been drawn in order to identify the best policies. However, the policy search itself has been performed by formalizing a search over a space of simple closed-form formulas as a multi-armed bandit problem.
This means that all policy candidates have to be created in an initial step at once and are subsequently evaluated.
The computational effort to follow this approach combinatorially explodes as soon as more complex solutions are required to solve more complicated control problems.

\section{Model-based Reinforcement Learning}
\label{section:rl}

Inspired by behaviorist psychology, RL is concerned with how software agents ought to take actions in an environment in order to maximize their received accumulated rewards.
In RL, the acting agent is not explicitly told which actions to implement.
Instead, the agent must learn the best action strategy from the observed environment's rewards in response to the agent's actions. 
Generally, such actions affect both the next reward and subsequent rewards~\citep{sutton:98}.

In RL formalism, at each discrete time step $t=0,1,2,\ldots$, the agent observes the system's state $\mathbf{s}_t \in \mathcal S$ and applies an action $\mathbf{a}_t \in \mathcal A$, where $\mathcal S$ is the state space and $\mathcal A$ is the action space. 
Depending on $\mathbf{s}_t$ and $\mathbf{a}_t$, the system transitions to the next state $\mathbf{s}_{t+1}$ and the agent receives a real-value reward $r_{t+1} \in \mathbb{R}$. 
In deterministic systems the state transition can be expressed as a function $g:\mathcal S \times \mathcal A \rightarrow \mathcal S$ with $g(\mathbf{s}_t,\mathbf{a}_t)=\mathbf{s}_{t+1}$.
The related reward is given by a reward function $r:\mathcal S \times \mathcal A \times \mathcal S \rightarrow \mathbb{R}$ with $r(\mathbf{s}_t,\mathbf{a}_t,\mathbf{s}_{t+1})=r_{t+1}$. 
Hence, the desired solution to an RL problem is a policy that maximizes the expected accumulated rewards.

In our proposed setup, the goal is to find the best policy $\pi$ among $\Pi$ the set of all possible equations which can be built from a pre-defined set of function building blocks, with respect to a certain maximum complexity.
For every state $\mathbf{s}_t$, the policy outputs an action, i.e., $\pi(\mathbf{s}_t)=\mathbf{a}_t$. 
The policy's performance, when starting from $\mathbf{s}_t$, is measured by the return $\mathcal{R}(\mathbf{s}_t,\pi)$, i.e., the accumulated future rewards obtained by executing the policy $\pi$. 
To account for increasing uncertainties when accumulating future rewards, the reward $r_{t+k}$ for $k$ future time steps is weighted by $\gamma^k$, where $\gamma\in[0,1]$.
Furthermore, adopting a common approach, we include only a finite number of $T>1$ future rewards in the return~\citep{sutton:98}, which is expressed as follows:
\begin{equation}\label{eq:return}
  \begin{aligned}
    \mathcal R (\mathbf{s}_t,\pi) & = \sum_{k=0}^{T-1}\gamma^kr(\mathbf{s}_{t+k},\pi(\mathbf{s}_{t+k}),\mathbf{s}_{t+k+1}), \\
    \textnormal{with}\quad \mathbf{s}_{t+k+1} & = g(\mathbf{s}_{t+k},\mathbf{a}_{t+k}).
  \end{aligned}
\end{equation}
Herein, we select the discount factor $\gamma$ such that, at the end of time horizon $T$, the last reward accounted for is weighted by $q\in[0,1]$, yielding $\gamma=q^{1/(T-1)}$.
The overall state-independent policy performance $\mathcal{F}(\pi)$ is obtained by averaging over all starting states $s_t \in S \subset\mathcal S$, using their respective probabilities $w_{\mathbf{s}_t}$ as weight factors. 
Thus, optimal solutions to the RL problem are policies $\pi$ with
\begin{equation}\label{eq:fitness_function}
    \hat{\pi} \in \argmax_{\pi \in \Pi}\mathcal{F}(\pi), \quad\text{with}\quad \mathcal{F}(\pi)=\frac{1}{\abs{S}}\sum_{\mathbf{s}_t\in S}w_{\mathbf{s}_t}\mathcal R (\mathbf{s}_t,\pi).
\end{equation}
In optimization terminology, the policy performance function $\mathcal{F}(\pi)$ is referred to as a fitness function.

For most real-world industrial control problems, the cost of executing a potentially bad policy is prohibitive. 
Therefore, in model-based RL~\citep{busoniu:10}, the state transition function $g$ is approximated using a model $\tilde g$, which can be a first principle model or can be created from previously gathered data. 
By substituting $\tilde{g}$ in place of the real system $g$ in \eqref{eq:return}, we obtain a model-based approximation $\tilde{\mathcal{F}}(\pi)$ of the true fitness function \eqref{eq:fitness_function}. 
In this study, we employ models based on NNs. 
However, the proposed method can be extended to other models, such as Bayesian NNs~\citep{depeweg:16} and Gaussian process models~\citep{rasmussen:06}.

\section{Genetic Programming}
\label{section:gp}

GP is a technique, which encodes computer programs as a set of genes. 
Applying a so-called genetic algorithm (GA) on these genes to modify (evolve) them drives the optimization of the population.
Generally, the space of solutions consists of computer programs, which perform well on predefined tasks~\citep{koza:92}.
Since we are interested in interpretable equations as RL policies, the genes in our setting include basic algebraic functions, as well as constant float numbers and state variables. 
Such basic algebraic functions can be depicted as function trees and stored efficiently in memory arrays.

\begin{figure}
	\centering
	\includegraphics[width=3.5in]{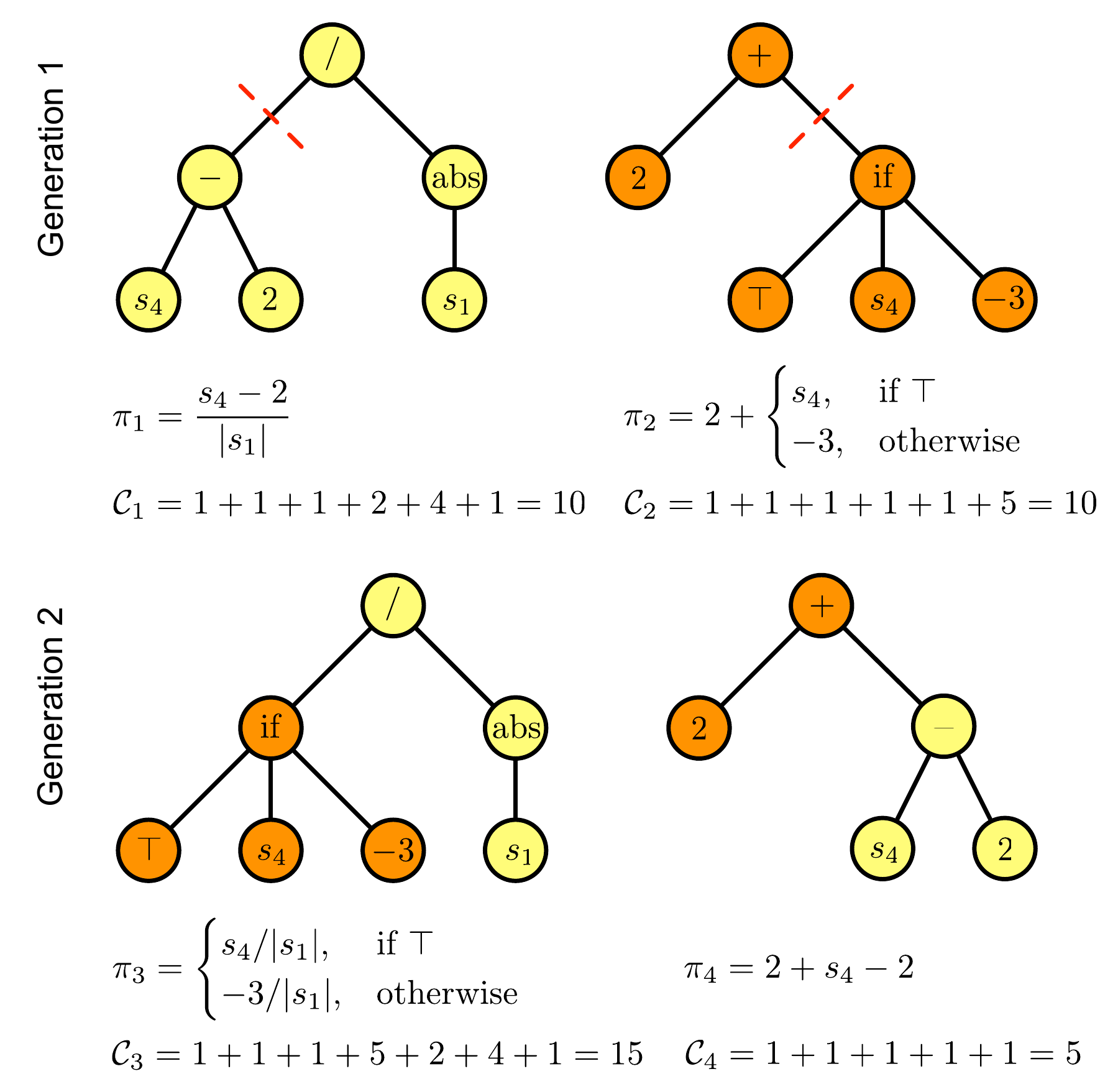}
	\caption{GP individuals as function trees.
		Depicted are four exemplary GP individuals $\pi_i$ from two consecutive generations with their respective complexity measures $\mathcal{C}_i$.
		Crossover cutting points are marked in the tree diagrams of policies $\pi_1$ and $\pi_2$.}
	\label{gp_trees}
\end{figure}

The GA drives the optimization by applying selection and reproduction on the populations.
The basis for both concepts is a fitness value $\mathcal{F}$ which represents the quality of performing the predefined task for each individual.
Selection means that only the best portion of the current generation will survive each iteration and continue existing in the next generation.
Analogous to biological sexual breeding, two individuals are selected for reproduction based on their fitness, and two offspring individuals are created by crossing their chromosomes.
Technically, this is realized by selecting compatible cutting points in the function trees and interchanging the subtrees beneath these cuts.
Subsequently, the two resulting individuals are introduced to the population of the next generation (Figure~\ref{gp_trees}).
Herein, we applied tournament selection~\citep{blickle:95} to select the individuals to be crossed.

In our experiments, it has shown to be advantageous to apply automatic equation cancellation  to a certain amount (defined by auto cancellation ratio $r_\text{a}$) of the best-performing individuals of one generation. 
For canceling, an algorithm is applied, which searches the chromosomes for easy-to-cancel subtree structures, calculates the result of the subtree and replaces it by this result. 
For example, if a subtree with a root node $+$ is found, whose children are two float terminal nodes $a$ and $b$, the subtree can be replaced by one float terminal node $c$, given by $c=a+b$. 
Similar cancelation rules can be derived for all functions in the function set. 
Since the tree depth is limited in our GP setup, this algorithm can reduce the complexity of substructures and even the whole individual, as well as generate space for potentially more important subtrees.

As a mutation operator, we adopted the so-called Gaussian mutator for float terminals, which is common in evolutionary algorithms~\citep{schwefel:81,schwefel:95}.
In each generation, a certain portion (according to terminal mutation ratio $r_\text{m}$) of the best-performing individuals for each complexity is selected.
Subsequently, these individuals are copied and their original float terminals $z$ are mutated by drawing replacement terminals $z'$ from a normal distribution $\mathcal{N}(z,0.1\abs{z})$.
If the performance of the best copy is superior to that of the original individual, it is added to the new population.
This strategy provides an option for conducting a local search in the policy search space, because the basic structure of the individual's genotype remains untouched.

Initially, the population is generated randomly, as well as a certain portion of each population every new generation (according to a new random individual ratio $r_\text{n}$).
A common strategy to randomly generate valid individuals is to apply the so-called grow method.
In our implementation, growing a chromosome is realized as follows:
\begin{enumerate}
	\item Randomly draw tree depth $d$ from $[d_\text{min},d_\text{max}]$
	\item select\_next\_gene($d$)
	\item Procedure: select\_next\_gene($d$)
	\begin{enumerate}
		\item \textbf{If} $d<1$\\
		Randomly draw gene $g$ from the set of terminals and variables\\
		\textbf{Else}\\
		Randomly draw gene $g$ from the set of functions
		\item Add $g$ to chromosome $c$
		\item Randomly select one leaf of $g$: $i$
		\item Build chromosome $c_i\leftarrow$select\_next\_gene($d-1$)
		\item Add $c_i$ to $c$
		\item \textbf{For} all leafs of node $j\neq i$
		\begin{enumerate}
			\item Randomly draw subtree depth $d_j$ from $[0,d-1]$
			\item Build chromosome $c_j\leftarrow$select\_next\_gene($d_j$)
			\item Add $c_j$ to $c$
		\end{enumerate}
		\item \textbf{Return} $c$
	\end{enumerate}
\end{enumerate}
Note that this algorithm enforces a broad variety of individuals, since it randomizes the length of each subtree (1. and 3.(f).i.), as well as the position of the biggest subtree (3.(c)).
Both properties save the GA from generating only small initial chromosomes, which eventually would result in an early convergence of the population.

The overall GA used in the experiments is given as follows:
\begin{enumerate}
	\item Randomly initialize the population of size $N$
	\item Determine fitness value of each individual (in parallel)
	\item Evolve next generation
	\begin{enumerate}
		\item Crossover (depending on crossover ratio $r_\text{c}$)
		\begin{enumerate}
			\item Select individuals by tournament selection
			\item Cross two tournament winners
			\item Add resulting individuals to new population
		\end{enumerate}
		\item Reproduction (depending on reproduction ratio $r_\text{r}$)
		\begin{enumerate}
			\item Select individuals by tournament selection
			\item Add tournament winner to new population
		\end{enumerate}
		\item Automatic cancelation and terminal adjustment (depending on auto cancel ratio $r_\text{a}$ and terminal adjustment ration $r_\text{m}$)
		\begin{enumerate}
			\item Apply automatic cancelation on all individuals
			\item Add canceled individuals according to $r_\text{a}$
			\item Select best individuals of old population
			\item Randomly mutate float terminals ($z'\sim z+0.1z\cdot\mathcal{N}(0,1)$) and create $N\cdot r_\text{a}$ adjusted individuals from each best
			\item Determine fitness value of each individual (in parallel)
			\item Add best adjusted individuals to new population according to $r_\text{m}$
		\end{enumerate}
		\item Fill new population with new randomly generated individuals (new individuals ratio  $r_\text{n}$)
		\item Determine fitness value of each individual (in parallel)
		\item \textbf{If} none of the stopping criteria is met
		\begin{enumerate}
			\item Go back to 3.
		\end{enumerate}
	\end{enumerate}
	\item \textbf{Return} best individual found so far for each complexity level
\end{enumerate}

Since in this work we are searching for interpretable solutions, a measure of complexity has to be established.
Measuring the complexity of an individual can generally be stated with respect to its genotype (structural) or with respect to its phenotype (functional)~\citep{le:16}.
Here, we decided to use a simple node-counting measuring strategy where different types of functions, variables and terminals are counted with different weightings.
Hence, the domain experts, to whom the RL policies might be delivered to, can pre-define which types of genes they are more likely to accept as interpretable compared to others.
In particular, we adopted the complexity weightings from Eureqa\footnote{\url{https://www.nutonian.com/products/eureqa}}, a commercial available software for symbolic regression~\citep{dubvcakova:11}.
Table~\ref{table:complexities} lists the weightings we applied in our experiments and Figure~\ref{gp_trees} gives four examples on how to calculate the respective complexities.

\begin{table}
\begin{center}
\begin{tabular}{ll} 
	\toprule
    Variables & 1\\
    Terminals & 1\\
    $+,-,\cdot$ & 1\\
    $/$ & 2\\
    $\land,\lor$ & 4\\
    $\text{tanh},\text{abs}$ & 4\\
    $\text{if}$ & 5\\
    \bottomrule	
\end{tabular}
\caption{GP complexities}
\label{table:complexities}
\end{center}
\end{table}

Table~\ref{table:setup_gp} gives an overview of the GP parameters and methods we used in the experiments below.
Note that we decided to employ rather big population sizes (up to 1000) in combination with a higher new individuals ratio ($r_\text{n}=0.3$), compared to other publications~\citep{koshiyama:14}.
By doing so, we empirically observed that GPRL converges to better solutions faster, compared to a setting with smaller population and more generations.
The latter setting very often converged to suboptimal solutions too early.
One reason for that might be the lack of diversity observable in smaller populations with a limited number of new individuals in each generation.
Furthermore, with the parallel computing options of today's computational resources, big populations can be evaluated in parallel, while many consecutive generations have to be evaluated sequentially.

\begin{table}
\begin{center}
\begin{tabular}{ll}  
	\toprule
    Individual representation & tree-based\\
    Initialization method & grow method\\
    Selection method & tournament selection (size=3)\\
    Terminal set & state variables, random float\\
    & numbers $z\sim[-20.0,20.0]$,\\
    & $\top, \bot$ \\
    Function set & $+,-,*,/,\land,\lor,\text{if},>,<,$\\
    &$\text{tanh},\text{abs}$ \\
    Maximal gene amount & 100\\
    Maximal tree depth & 5\\
    Maximal complexity & 100\\
    Ratios for new generation & crossover $r_\text{c}=0.45$\\
    & reproduction $r_\text{r}=0.05$\\
    & auto cancel  $r_\text{a}=0.1$\\
    & terminal mutation $r_\text{m}=0.1$\\
    & new random individuals\\ 
    & $r_\text{n}=0.3$\\
    \midrule
    Population/generations/& MC: 100/1,000/1,000\\
    training states & CPB: 1,000/1,000/1,000\\
    for GPRL & IB: 1,000/1,000/100\\
    \midrule
    Population/generations/ & MC: 1,000/1,000/70,000\\
    samples& CPB: 10,000/1,000/70,000\\
	for symbolic regression & IB: 10,000/1,000/100,000\\
	\bottomrule	
\end{tabular}
\caption{GP parameters. 
	Note, that the population and generation numbers have been determined empirically.
	The generation numbers are chosen such that after these points no substantially better individuals have been found throughout the population.}
\label{table:setup_gp}
\end{center}
\end{table}

\section{Genetic Programming Reinforcement Learning}
\label{section:gprl}

The basis for the proposed GPRL approach is a data set $\mathcal{D}$ that contains state transition samples gathered from the dynamics of a real system. 
These samples are represented by tuples $(\mathbf{s},\mathbf{a},\mathbf{s}',r)$, 
where, in state $\mathbf{s}$, action $\mathbf{a}$ was applied and resulted in state transition to $\mathbf{s}'$.
Subsequently, this transition yielded a real value reward $r$.
Note that generally $\mathcal{D}$ can be generated using any (even a random) policy prior to policy training as long as sufficient exploration is involved~\citep{sutton:98}.

In a first step, we generate world models $\tilde g$ with inputs $(\mathbf{s},\mathbf{a})$ to predict $\mathbf{s}'$, using data set $\mathcal{D}$. 
To yield better approximative quality, we observed that for many problems it is advantageous to learn the differences between $\mathbf{s}$ and $\mathbf{s'}$ and to train a single model per state variable separately:
\begin{align*}
\Delta s'_1 & = \tilde g_{s_1}(s_1,s_2,\dots,s_m,\mathbf{a}),\\
\Delta s'_2 & = \tilde g_{s_2}(s_1,s_2,\dots,s_m,\mathbf{a}),\\
\dots\\
\Delta s'_m & = \tilde g_{s_m}(s_1,s_2,\dots,s_m,\mathbf{a}).
\end{align*}
Hence, the resulting state is calculated according to $\mathbf{s}'=(s_1+\Delta s'_1,s_2+\Delta s'_2,\dots,s_m+\Delta s'_m)$.
Note that the reward is also given in data set $\mathcal{D}$; thus, the reward function can also be approximated using $r=\tilde r(\mathbf{s},\mathbf{a},\mathbf{s}')$.

The interpretable policies we are generating applying our GPRL approach in Section~\ref{section:experiments} are basic algebraic equations.
Given that GPRL is able to find rather short (non-complex) equations, we expect to reveal substantial knowledge about underlying coherencies between available state variables and well-performing control policies with respect to a certain RL problem.
To rate the quality of each policy candidate a fitness value has to be provided for the GP algorithm to advance.
For our GPRL approach, the fitness $\mathcal{\tilde F}$ of each individual is calculated by generating trajectories using the world model $\tilde g$ starting from a fixed set of initial benchmark states (Section~\ref{section:rl}).

The performance of GPRL is compared to a rather straightforward approach, which utilizes GP to conduct symbolic regression on a data set $\hat{\mathcal{D}}$ generated by a well-performing but non-interpretable RL policy $\hat{\pi}$.
$\hat{\mathcal{D}}$ contains tuples $(\mathbf{s},\hat{\mathbf{a}})$, where $\hat{\mathbf{a}}$ are the generated actions of policy $\hat{\pi}$ on state $\mathbf{s}$.
The states originate from trajectories created by policy $\hat{\pi}$ on world model $\tilde g$.
One might think that given an adequate policy of any form and using GP to mimic this policy by means of some regression error with respect to $\hat{\mathbf{a}}$, could also yield successful interpretable RL policies.
However, our results clearly indicate that this strategy is only successful for rather small and simple problems and produces highly non-stable and unsatisfactory results for more complex tasks.

In our experiments, the well-performing but non-interpretable RL policy $\hat{\pi}$ we used to generate data set $\hat{\mathcal{D}}$ is a NN policy.
To yield comparable results we always trained the weights of this policy by model-based RL on the very same world models as applied for GPRL.
Note, that usually we expect $\hat{\pi}$ to yield higher fitness values during training, since it is able to utilize significantly more degrees of freedom to compute an optimal state action mapping than basic algebraic equations found by GPRL.

Note that we use NNs as world models $\tilde g$ for the GPRL experiments (not to be confused with NN policy $\hat{\pi}$).
In many real-world industrial problem domains, i.e., continuous and rather smooth system dynamics, NNs are known to serve as adequate world models with excellent generalization properties. 
Given a batch of previously generated transition samples, the NN training process is known to be data-efficient.
Moreover, the training errors are excellent indicators of how well the model will perform in model-based RL training.
Nevertheless, for other problem domains, alternative types of world models might be preferable. 
For example, Gaussian processes~\citep{rasmussen:06} provide a good approximation of the mean of the target value, and  this technique indicates the level of confidence about this prediction, which may be of value for stochastic system dynamics.
Another alternative modeling technique is the use of regression trees~\citep{breiman:84}.
While typically lacking data efficiency, regression tree predictions are less affected by nonlinearities perceived by system dynamics because they do not rely on a closed-form functional approximation.

\begin{figure*}
	\centering
	\includegraphics[width=5in]{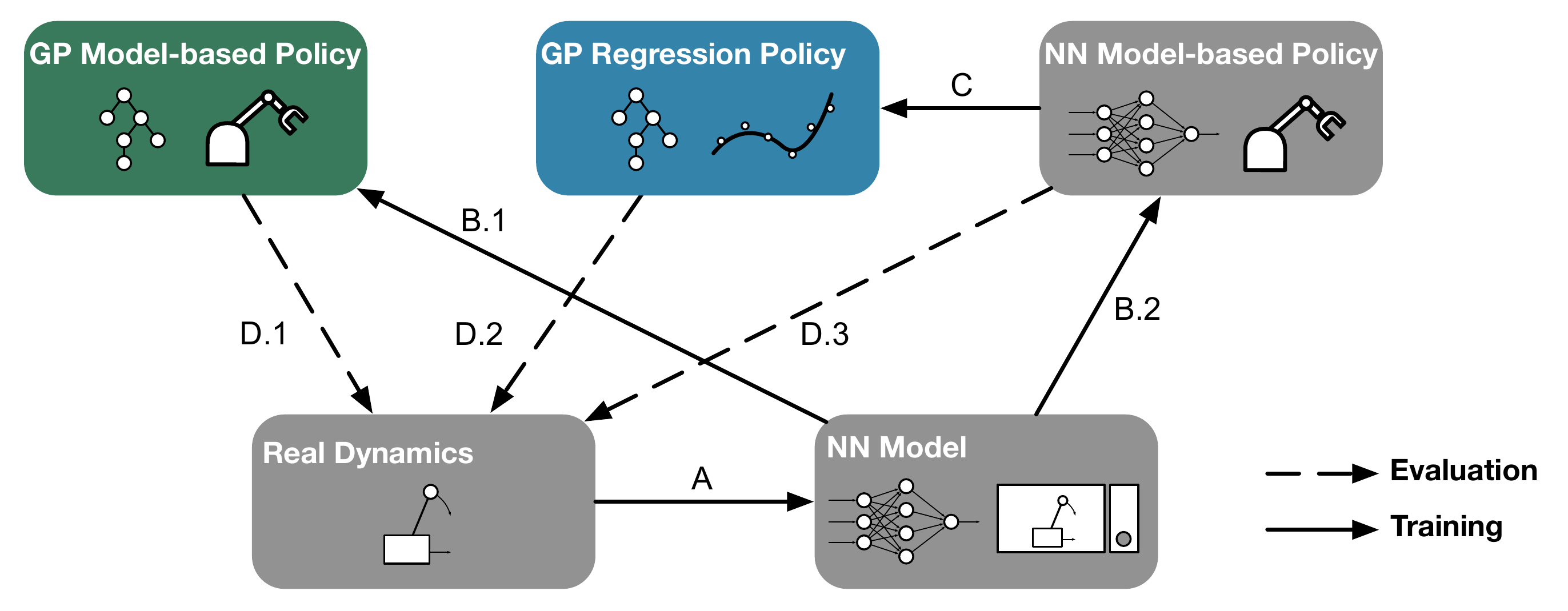}
	\caption{The proposed GPRL approach and its integration in the experimental setup.
		The world model is the result of supervised ML regression on data originated from the real dynamics (A).
		GPRL generates a GP model-based policy by training on this world model (B.1), which is an NN model in our experimental setup.
		Similarly, the NN policy is trained in a model-based manner by utilizing the same NN model (B.2).
		In contrast to both other policy training approaches, the GP regression policy mimics an already existing policy by learning to minimize an error with respect to the existing policy's action (C).
		All of the policies are finally evaluated by comparing their performance on the real dynamics (D.1-D.3).}
	\label{schema}
\end{figure*}

Figure~\ref{schema} gives an overview of the relationships between the different policy implementations, i.e., GPRL, GP for symbolic regression, and NN policy, and the environment instances, i.e., NN system model and the real dynamics, used for training and evaluation, respectively.

\section{Experiments}
\label{section:experiments}

\subsection{Mountain Car}

\begin{figure}
	\centering
	\includegraphics[width=2.55in]{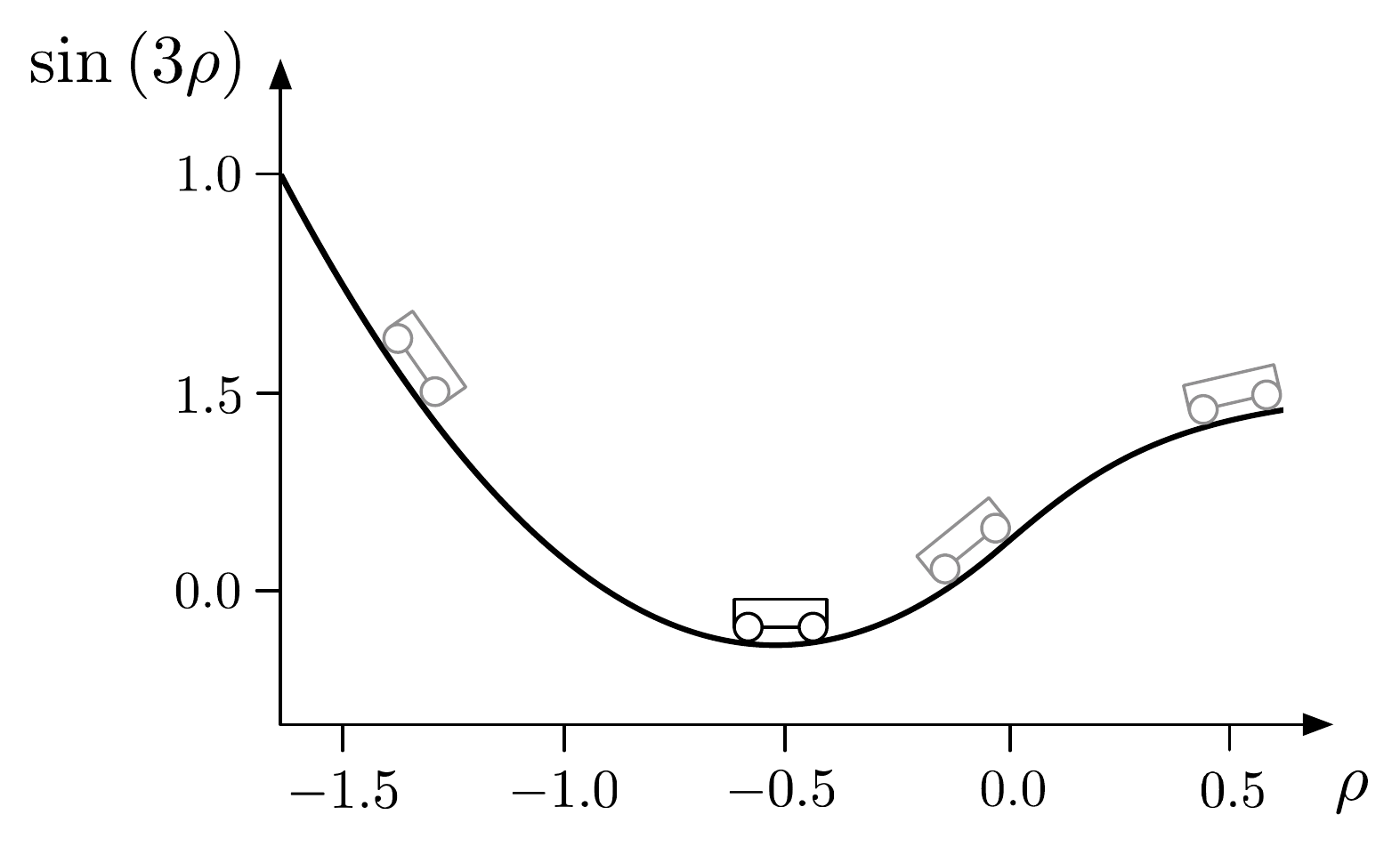}
	\caption{
		Mountain car benchmark.
		The task is to first build up momentum by driving to the left in order to subsequently reach the top of the hill on the right at $\rho=0.6$.
	}
	\label{mountain_car}
\end{figure}

In the MC benchmark, an underpowered car must be driven to the top of a hill (Figure~\ref{mountain_car}) \citep{moore:90}. 
This is achieved by building sufficient potential energy by first driving in the direction opposite to the final direction. 
The system is fully described by the two-dimensional state space $\mathbf{s}=(\rho,\dot{\rho})$ representing the cars position $\rho$ and velocity $\dot{\rho}$.

We conducted MC experiments using the freely available $CLS^2$ software ('clsquare')\footnote{\url{http://ml.informatik.uni-freiburg.de/research/clsquare}}, which is an RL benchmark system that applies the Runge-Kutta fourth-order method to approximate closed loop dynamics.
The task for the RL agent is to find a policy producing action sequence $a_t,a_{t+1},a_{t+2},\ldots\in[-1,1]$ that drive the car up the hill, which is achieved when reaching position $\rho\geq0.6$.
 
The agent receives a reward of 
\begin{equation}
	r(\mathbf{s}')=
	\begin{cases}
		0, & \text{if }\rho'\geq0.6,\\
		-1, & \text{otherwise},
	\end{cases}
\end{equation}
subsequent to each state transition $\mathbf{s}'=g(\mathbf{s},a)$. 
When the car reaches the goal position, i.e., $\rho\geq0.6$, its position becomes fixed and the agent receives the maximum reward in each following time step regardless of the applied actions.

\subsection{Cart-pole Balancing}

\begin{figure}
	\centering
	\includegraphics[width=1.8in]{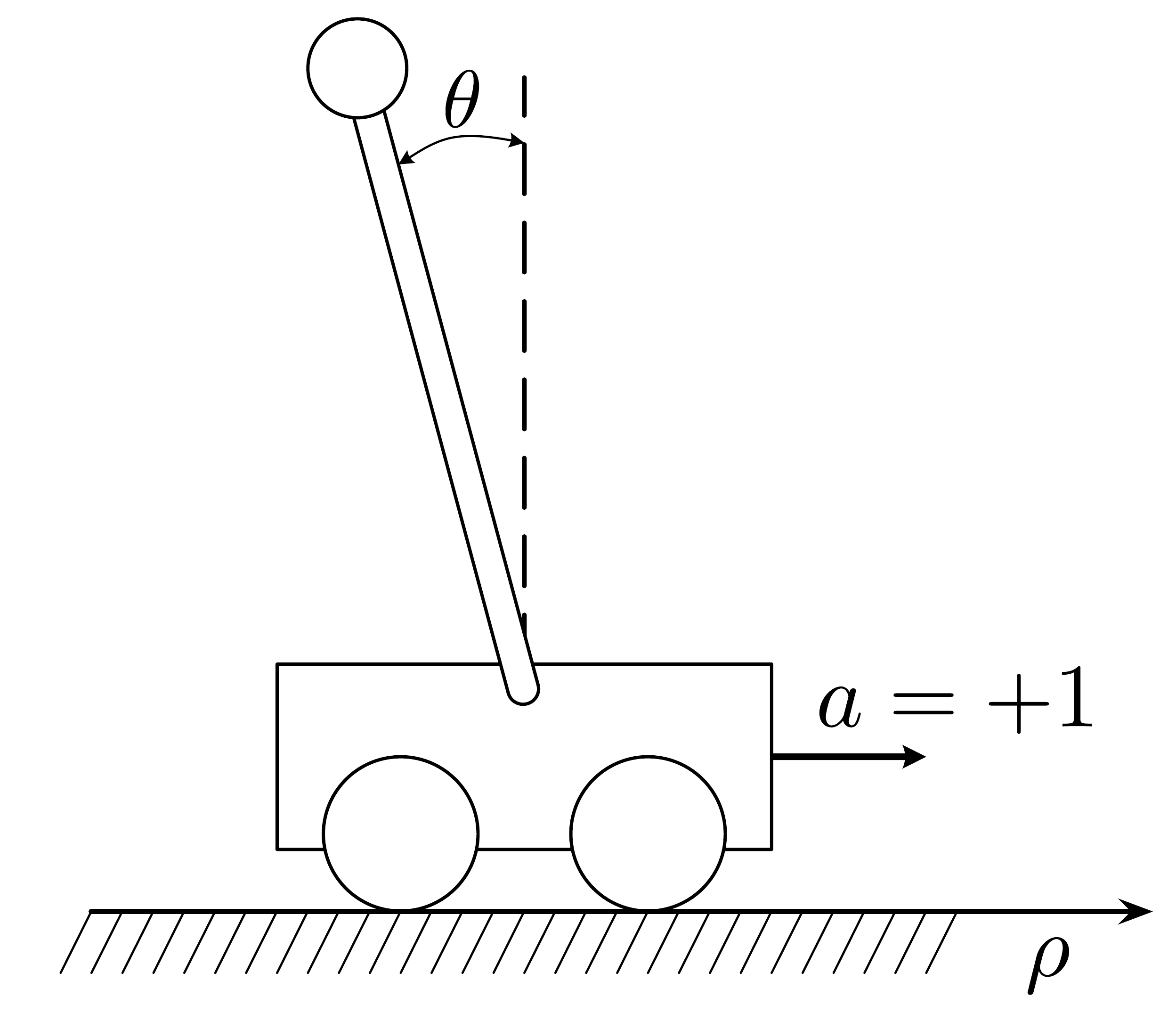}
	\caption{
		Cart-pole benchmark.
		The task is to balance the pole around $\theta=0$ while moving the cart to position $\rho=0$ by applying positive or negative force to the cart.
	}
	\label{cart_pole}
\end{figure}

The CPB experiments described in the following section were also conducted using the $CLS^2$ software.
The objective of the CPB benchmark is to apply forces to a cart moving on a one-dimensional track to keep a pole hinged to the cart in an upright position (Figure~\ref{cart_pole}). 
Here, the four Markov state variables are the pole angle $\theta$, the pole angular velocity $\dot\theta$, the cart position $\rho$, and the cart velocity $\dot\rho$. 
These variables describe the Markov state completely, i.e., no additional information about the system's past behavior is required. 
The task for the RL agent is to find a sequence of force actions $a_t,a_{t+1},a_{t+2},\ldots$  that prevent the pole from falling over~\citep{fantoni:02}.

In the CPB task, the angle of the pole and the cart's position are restricted to intervals of $[-0.7,0.7]$ and $[-2.4,2.4]$ respectively. 
Once the cart has left the restricted area, the episode is considered a failure and the system remains in the failure state for the rest of the episode. 
The RL policy can apply force actions on the cart from $-10$~N to $+10$~N in time intervals of $0.025$~s.

The reward function for the balancing problem is given as follows:
\begin{equation}
	r(\mathbf{s}')=
	\begin{cases}
		0.0, & \text{if }\abs{\theta'}<0.25\\& \text{and } \abs{\rho'}<0.5,\\
		-1.0, & \text{if }\abs{\theta'}>0.7\\& \text{or } \abs{\rho'}>2.4,\\
		-0.1, & \text{otherwise}.
	\end{cases}
\end{equation}
Based on this reward function, the primary goal of the policy is to avoid reaching the failure state. 
The secondary goal is to drive the system to the goal state region where $r=0$ and keep it there for the rest of the episode.

\subsection{Industrial Benchmark}

The IB\footnote{\url{http://github.com/siemens/industrialbenchmark}} was designed to emulate several challenging aspects eminent in many industrial applications~\citep{hein:17a,hein:17b,hein:18}. 
It is not designed to be an approximation of any specific real-world system, but to pose a comparable hardness and complexity found in many industrial applications.

State and action spaces are continuous. 
Moreover, the state space is high-dimensional and only partially observable. 
The actions consist of three continuous components and affect three control inputs. 
Moreover, the IB includes stochastic and delayed effects. 
The optimization task is multi-criterial in the sense that there are two reward components that show opposite dependencies on the actions. 
The dynamical behavior is heteroscedastic with state-dependent observation noise and state-dependent probability distributions, based on latent variables. 
Furthermore, it depends on an external driver that cannot be influenced by the actions.

At any time step $t$ the RL agent can influence the IB via actions $\mathbf{a}_t$ that are three dimensional vectors in $[-1,1]^3$. 
Each action can be interpreted as three proposed changes to three observable state control variables. 
Those variables are: velocity $v$, gain $g$, and shift $h$. 
Each variable is limited to $[0,100]$ and calculated as follows:
\begin{align*}
	\mathbf{a}_t & = \left(\Delta v_t,\Delta g_t, \Delta h_t\right),\\
	v_{t+1} & =\max\left(0,\min\left(100,v_t+d^\text{v}\Delta v_t\right)\right),\\
	g_{t+1} & =\max\left(0,\min\left(100,g_t+d^\text{g}\Delta g_t\right)\right),\\
	h_{t+1} & =\max\left(0,\min\left(100,h_t+d^\text{h}\Delta h_t\right)\right),
\end{align*}
with scaling factors $d^\text{v}=1$, $d^\text{g}=10$, and $d^\text{h}=5.75$.

After applying the action $\mathbf{a}_t$, the environment transitions to the next time step $t+1$, yielding the internal state $\mathbf{s}_{t+1}$. 
State $\mathbf{s}_t$ and successor state $\mathbf{s}_{t+1}$ are the Markovian states of the environment, which are only partially observable by the agent. 
In addition to the three control variables velocity $v$, gain $g$, and shift $h$, a setpoint $p$ is applied to the system. 
Setpoint $p$ simulates an external force like the demanded load in a power plant or the wind speed actuating a wind turbine, which cannot be controlled by the agent, but still has a major influence on the system dynamics. 
Depending on the setpoint $p_t$ and the choice of control values $\mathbf{a}_t$, the system suffers from detrimental fatigue $f_t$ and consumes resources such as power, fuel, etc., represented by consumption $c_t$. 
In each time step, the IB generates output values for $c_{t+1}$ and $f_{t+1}$, which are part of the internal state $\mathbf{s}_{t+1}$.
The reward is solely determined by $\mathbf{s}_{t+1}$ as follows:
\begin{equation}
	r_{t+1}=-c_{t+1}-3f_{t+1}.
\end{equation}

Note that the complete Markov state $\mathbf{s}$ of the IB remains unobservable. 
Only an observation vector $\mathbf{o}\subset\mathbf{s}$ consisting of:
\begin{itemize}
	\itemsep0em
	\item the current control variables velocity $v_t$, gain $g_t$, and shift $h_t$,
	\item the external driver set point $p_t$, and
	\item the reward relevant variables consumption $c_t$ and fatigue $f_t$,
\end{itemize}
can be observed externally.

In Section~\ref{section:rl} the optimization task in model-based RL is described as working on the Markovian state $\mathbf{s}$ of the system dynamics. 
Since this state is not observable in the IB environment $\mathbf{s}_t$ is approximated by a sufficient amount of historic observations $\left(\,\mathbf{o}_{t-H},\mathbf{o}_{t-H+1},\ldots,\mathbf{o}_{t}\right)$ with time horizon $H$. 
Given a system model $g\left(\,\mathbf{o}_{t-H},\mathbf{o}_{t-H+1},\ldots,\mathbf{o}_{t},\mathbf{a}_t\right)=\left(\mathbf{o}_{t+1},r_{t+1}\right)$ with $H=30$ an adequate prediction performance could be achieved during IB experiments. 
Note that observation size $\abs{\mathbf{o}}=6$ in combination with time horizon $H=30$ results in a 180-dimensional approximation vector of the Markovian state.

\subsection{Neural Network World Models}

The model-based policy training has been performed on NN world models, which yielded approximative fitness functions $\tilde {\mathcal{F}}(\mathbf{x})$ (Section~\ref{section:rl}).
For these experiments, we created one NN for each state variable. 
Prior to training, the respective data sets were split into blocks of 80\%, 10\%, and 10\% (training, validation and generalization sets, respectively).
While the weight updates during training were computed by utilizing the training sets, the weights that performed best given the validation sets were used as training results.
Finally, those weights were evaluated using the generalization sets to rate the overall approximation quality on unseen data.

The MC NNs were trained with data set $\mathcal{D}_{\text{MC}}$ containing tuples $(\mathbf{s},a,g(\mathbf{s},a),r)$ from trajectories generated by applying random actions on the benchmark dynamics. 
The start states for these trajectories were uniformly sampled as $\mathbf{s}=(\rho,\dot{\rho})\in[-1.2,0.6]\times\{0\}$, i.e., at a random position on the track with zero velocity. 
$\mathcal{D}_{\text{MC}}$ contains 10,000 transition samples.
The following three NNs were trained to approximate the MC task:
\begin{align*}
\Delta \rho_{t+1} & = \tilde g_{\rho}(\rho_t,{\dot\rho}_t,a_t),\\
\Delta \dot\rho_{t+1} & = \tilde g_{\dot\rho}(\rho_t,{\dot\rho}_t,a_t),\\
r_{t+1} & = \tilde r(\mathbf{s}_{t},a_t,\mathbf{s}_{t+1}),\\
\text{with }\mathbf{s}_{t+1} &= (\rho_t+\Delta \rho_{t+1},\dot{\rho}_t+\Delta \dot\rho_{t+1}).
\end{align*}

Similarly, for the CPB dynamic model state $\mathbf{s}_{t}=(\theta_t,{\dot\theta}_t,\rho_t,{\dot\rho}_t)$ we created the following four networks:
\begin{align*}
\Delta \theta_{t+1} & = \tilde g_{\theta}(\theta_t,{\dot\theta}_t,\rho_t,{\dot\rho}_t,a_t)\\
\Delta \dot\theta_{t+1} & = \tilde g_{\dot\theta}(\theta_t,{\dot\theta}_t,\rho_t,{\dot\rho}_t,a_t)\\
\Delta \rho_{t+1} & = \tilde g_{\rho}(\theta_t,{\dot\theta}_t,\rho_t,{\dot\rho}_t,a_t)\\
\Delta \dot\rho_{t+1} & = \tilde g_{\dot\rho}(\theta_t,{\dot\theta}_t,\rho_t,{\dot\rho}_t,a_t).
\end{align*}
An approximation of the next state is given by the following formula:
\begin{equation}
\mathbf{s}_{t+1}=(\theta_t+\Delta \theta_{t+1},{\dot\theta}_t+\Delta \dot\theta_{t+1},\rho_t+\Delta \rho_{t+1},{\dot\rho}_t+\Delta \dot\rho_{t+1}).
\end{equation}
The result of this formula can subsequently be used to approximate the state transition's reward by
\begin{equation}
r_{t+1} = \tilde r(\mathbf{s}_{t},a_t,\mathbf{s}_{t+1}).
\end{equation}
For the training set $\mathcal{D}_{\text{CPB}}$ of the CPB benchmark, the samples originate from trajectories of 100 state transitions generated by a random walk on the benchmark dynamics.
The start states $(\theta,\dot{\theta},\rho,\dot{\rho})$ for these trajectories were sampled uniformly from $[-0.7,0.7]\times\{0\}\times[-2.4,2.4]\times \{0\}$.
$\mathcal{D}_{\text{CPB}}$ contains 10,000 transition samples.

The experiments for MC and CPB were conducted with a network complexity of three hidden layers with 10 hidden neurons each and rectifier  activation functions.
For training, we used the Vario-Eta algorithm~\citep{montavon:12}.

For the IB benchmark two recurrent neural networks (RNNs) have been trained:
\begin{align*}
	c_{t+1} & = \tilde g_{c}(\mathbf{o}_t\setminus\{c_t\},\mathbf{a}_t),\\
	f_{t+1} & = \tilde g_{f}(\mathbf{o}_t,\mathbf{a}_t),\\
	r_{t+1} & = -c_{t+1}-3f_{t+1}.
\end{align*}
Note that the Markov decision process extraction topology~\citep{duell:12} of the RNNs that we applied here is well-suited for partially observable problems like the IB.
Detailed information on this topology, other design decisions, the training data, and the training process have been previously published by \cite{hein:17b}.

\section{Results}
\label{section:results}

\subsection{Mountain Car}

We conducted the MC experiments using a time horizon $T$ of 200 and a discount vector $\gamma$ of 0.985 ($q=0.05$).
For the MC experiments, a non-interpretable NN policy with fitness value $\mathcal{F}=-41.0$ (equivalent to a penalty value of 41.0) has been trained prior to the GP experiments.
A policy with this fitness value is capable of driving the car to the top of the hill from every state in the test set.
The NN policy has two hidden layers with $\tanh$ activation function and 10 hidden neurons on each layer.
Note that recreating such a policy with function trees as considered for our GPRL approach would result in a complexity value of 1581.

\begin{figure}
	\centering
	\subfloat[MC]{
		\includegraphics[trim=0 0 0 0,width=3in]{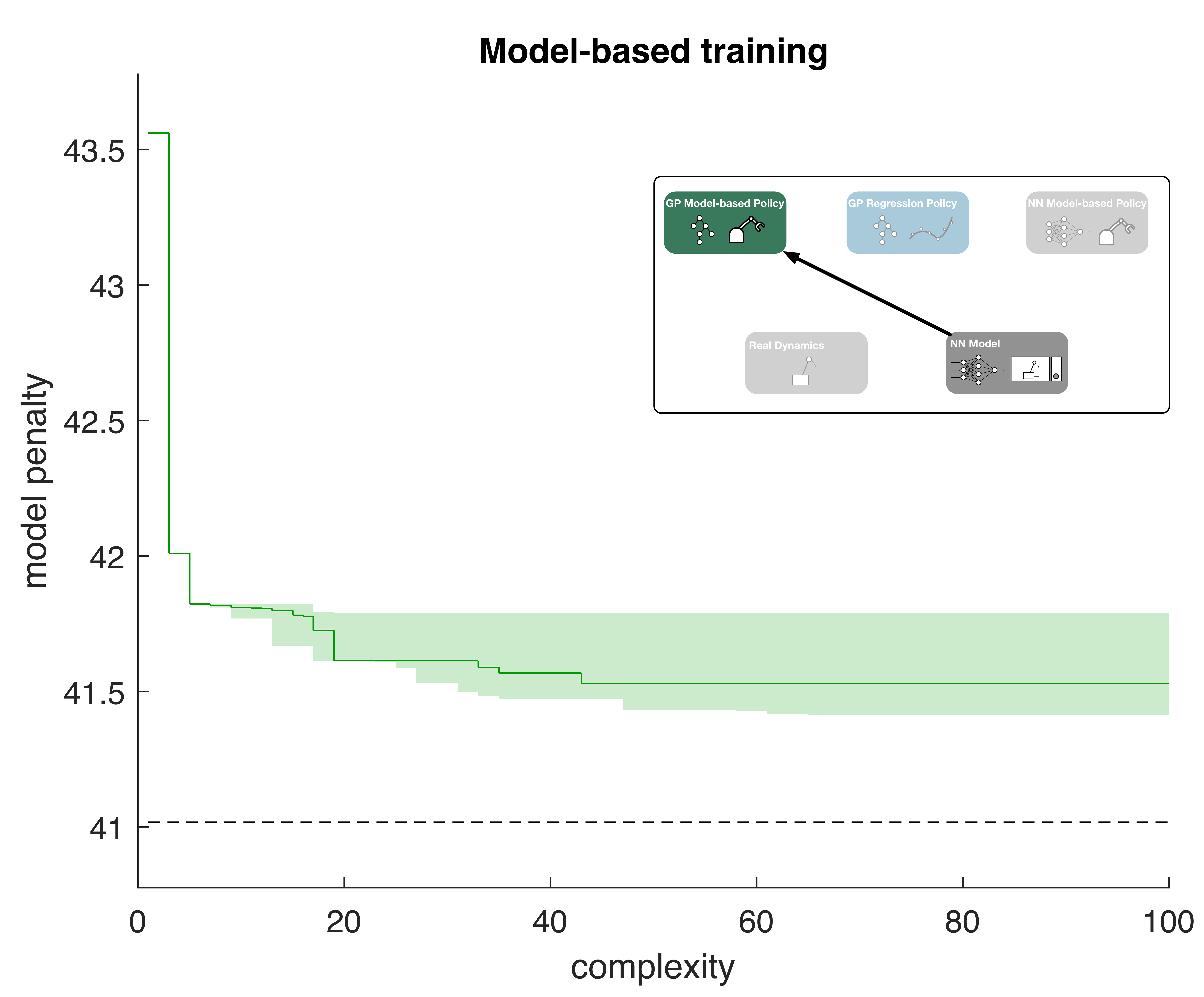}
		\label{mc_pareto_model}
	}\\
	\subfloat[CPB]{
		\includegraphics[trim=0 0 0 90,clip,width=3in]{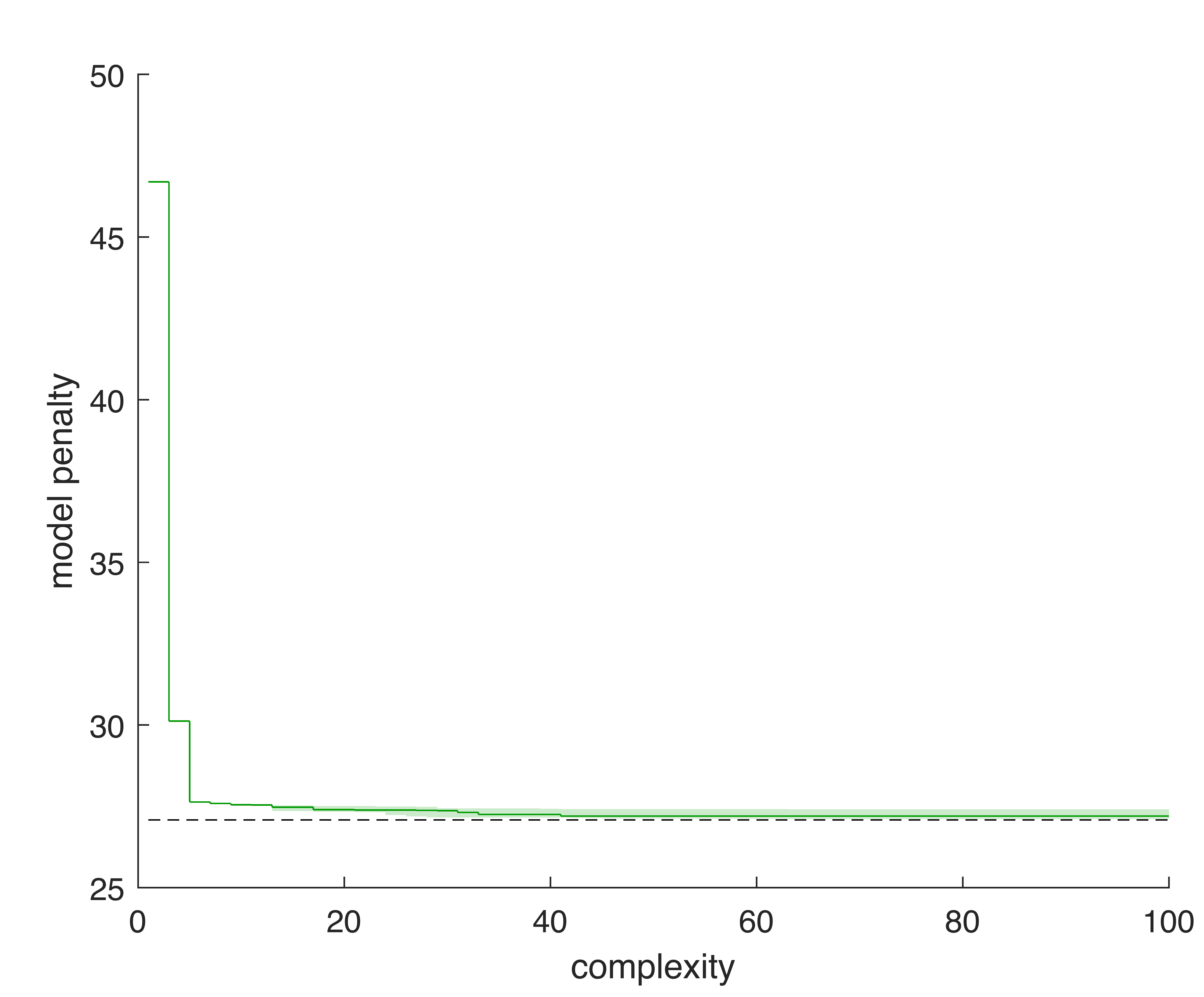}
		\label{cpb_pareto_model}
	}\\
	\subfloat[IB]{
		\includegraphics[trim=0 0 0 50,clip,width=3in]{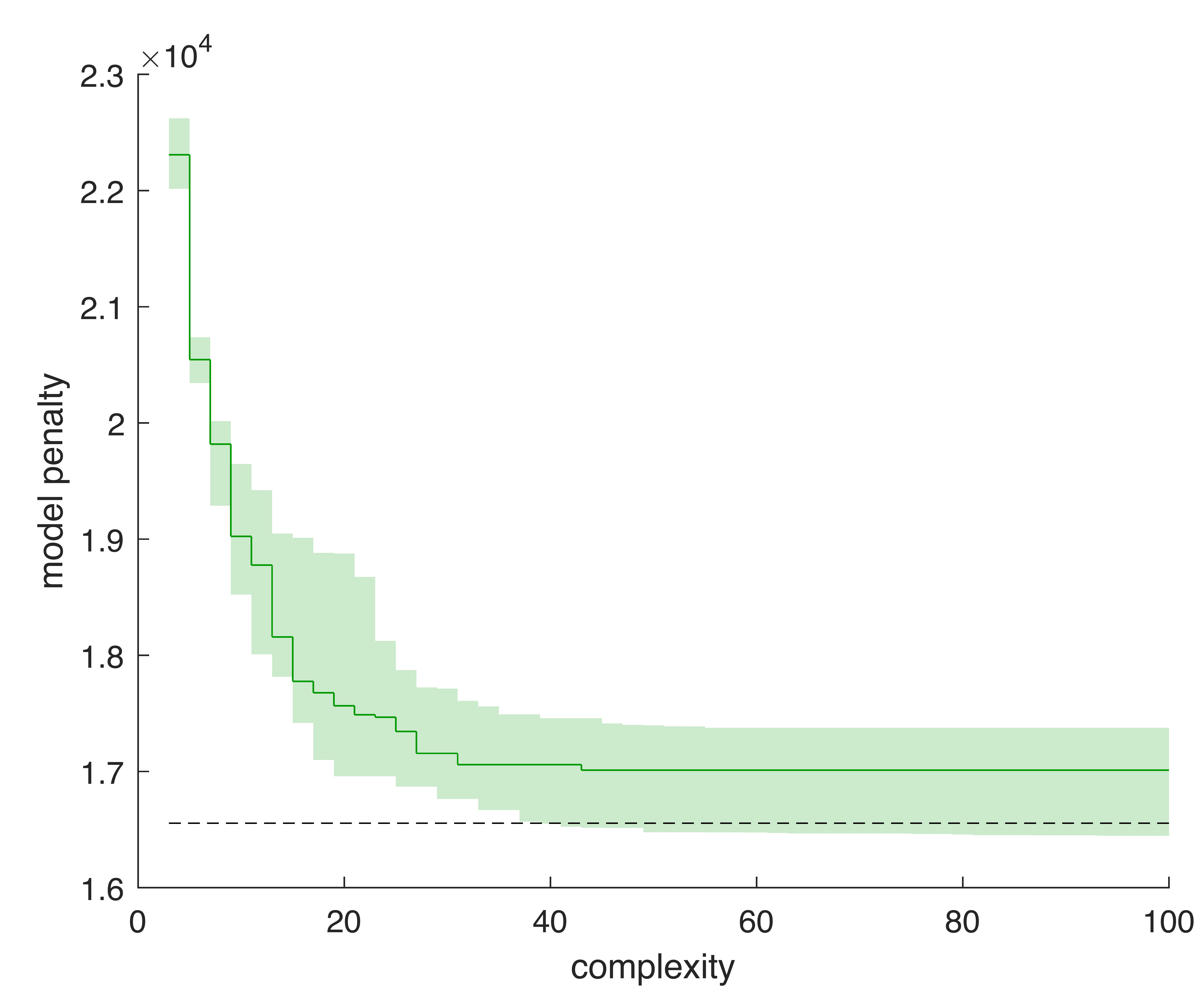}
		\label{ib_pareto_model}
	}
	\caption{Pareto fronts from ten model-based GPRL trainings.
		Depicted is the median (green line) together with the minimum and maximum (semi-transparent green area) Pareto front penalty from all experiments.
		The dashed line depicts the performance baseline of the NN policy on the NN model.}
	\label{pareto_model}
\end{figure}

The ten GPRL runs learned interpretable policies with a median model penalty of 41.8 for complexities $\geq5$ (Figure~\ref{mc_pareto_model}).
However, even the policies of complexity 1 managed to drive the car to the top of the hill, by simply applying the force along the direction of the car's velocity, i.e., $\pi(\rho,\dot{\rho})=\dot{\rho}$.
Though, policies with lower penalty managed to reach the top of the hill in fewer time steps.

\begin{figure*}
	\centering
	\includegraphics[width=7in]{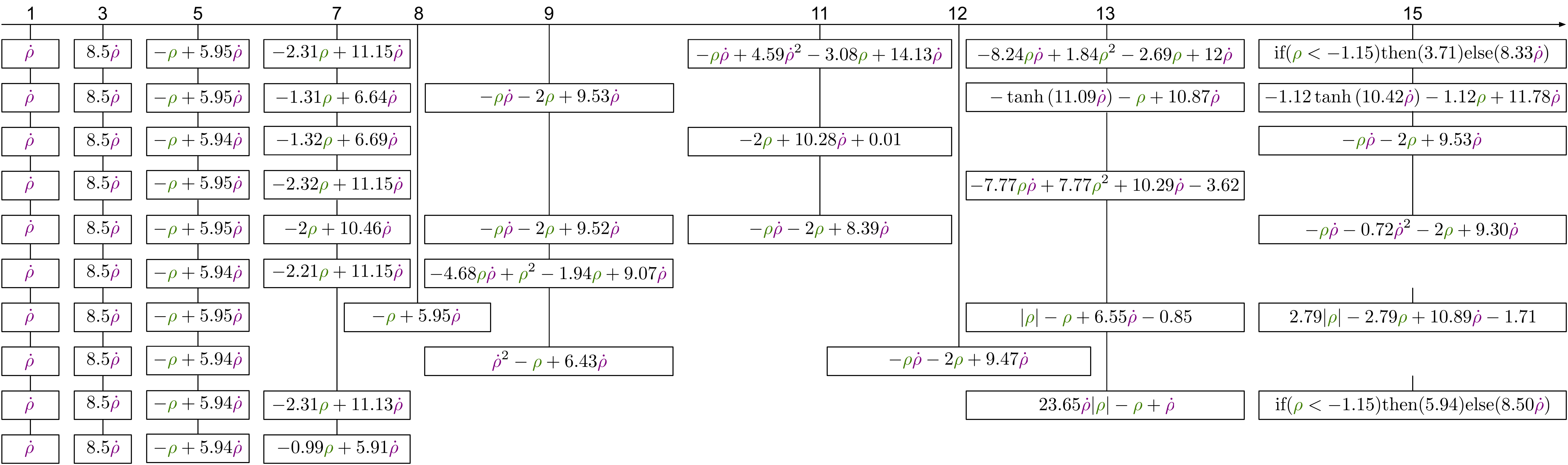}
	\caption{Interpretable MC policy results by our GPRL approach for complexities 1-15}
	\label{mc_solutions}
\end{figure*}

The resulting GPRL Pareto front individuals from complexity 1 to 15 are shown in Figure~\ref{mc_solutions}.

\begin{figure}
\centering
\subfloat[MC]{
	\includegraphics[trim=0 0 0 0,width=3in]{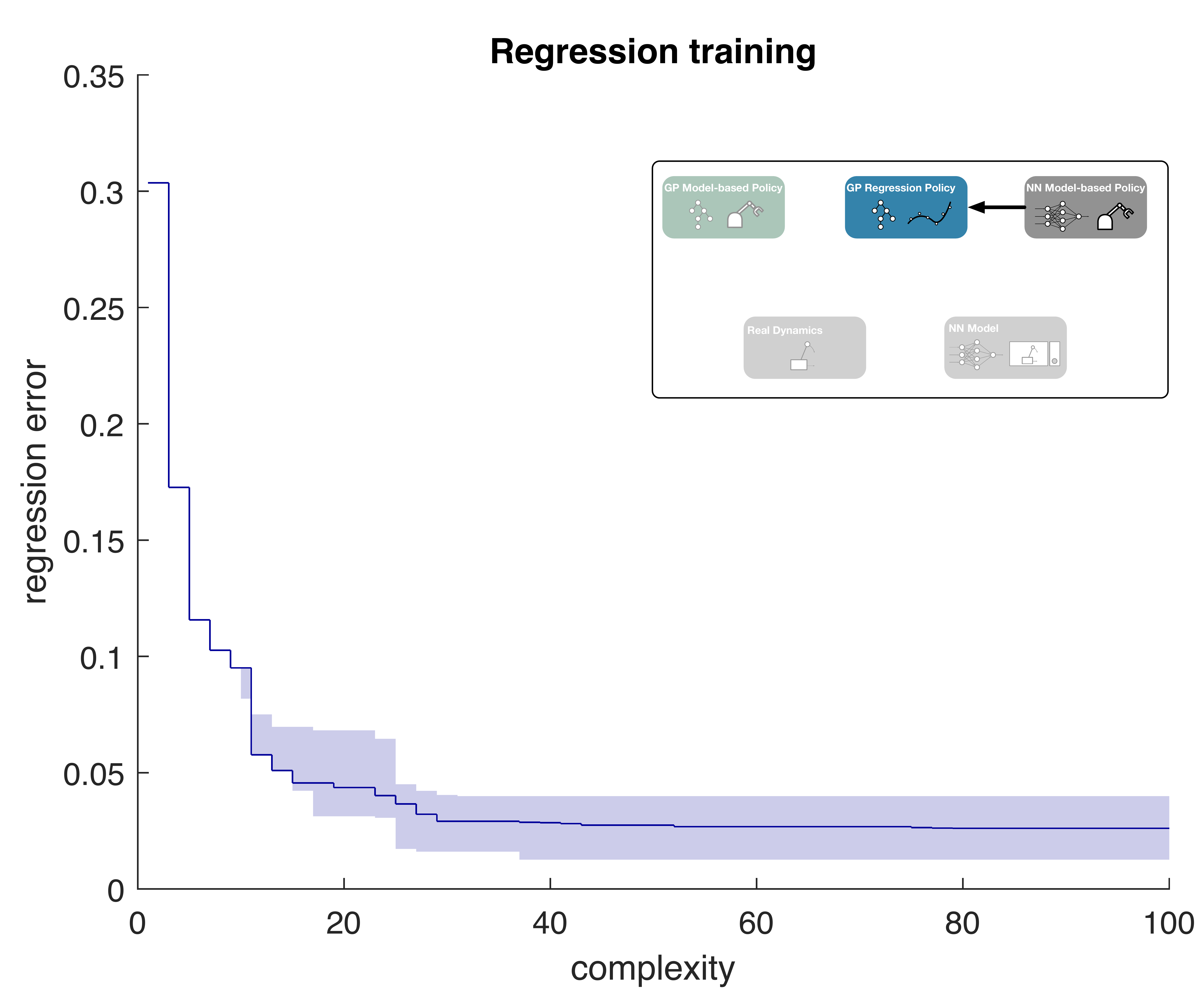}
	\label{mc_pareto_fit}
}\\
\subfloat[CPB]{
	\includegraphics[trim=0 0 0 90,clip,width=3in]{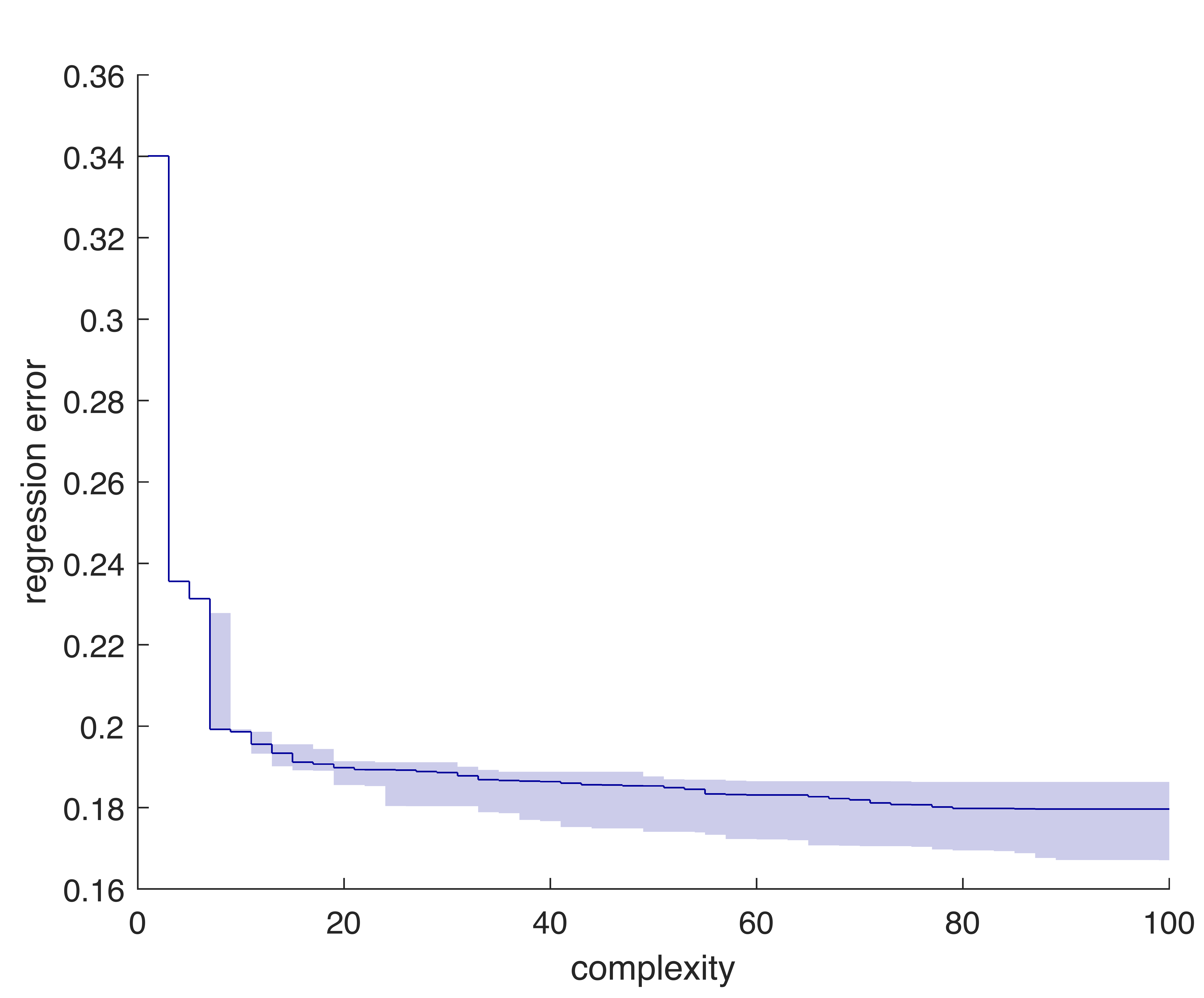}
	\label{cpb_pareto_fit}
}\\
\subfloat[IB]{
	\includegraphics[trim=0 0 0 50,clip,width=3in]{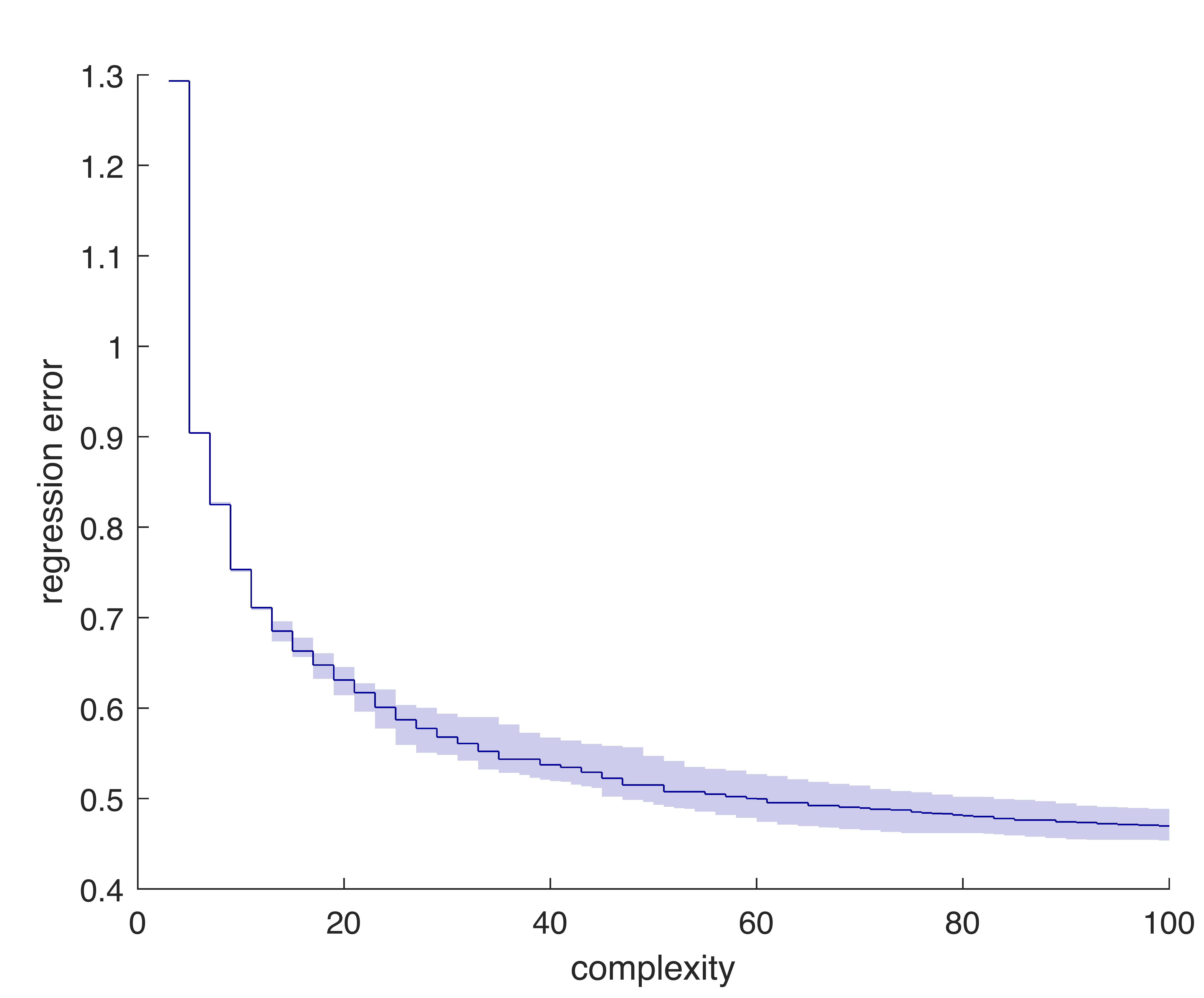}
	\label{ib_pareto_fit}
}
\caption{Pareto fronts from ten symbolic regression GP trainings.
	Depicted is the median (blue line) together with the minimum and maximum (semi-transparent blue area) Pareto front regression error from all experiments.}
\label{pareto_fit}
\end{figure}

Performing ten symbolic regression runs on the non-interpretable NN policy yielded interpretable policies with a median of the regression errors of 0.028 at best (Figure~\ref{mc_pareto_fit}).
This rather low regression error suggests a good performance on imitating the NN policy. 

\begin{figure}
	\centering
	\subfloat[MC]{
		\includegraphics[trim=0 0 0 0,width=3in]{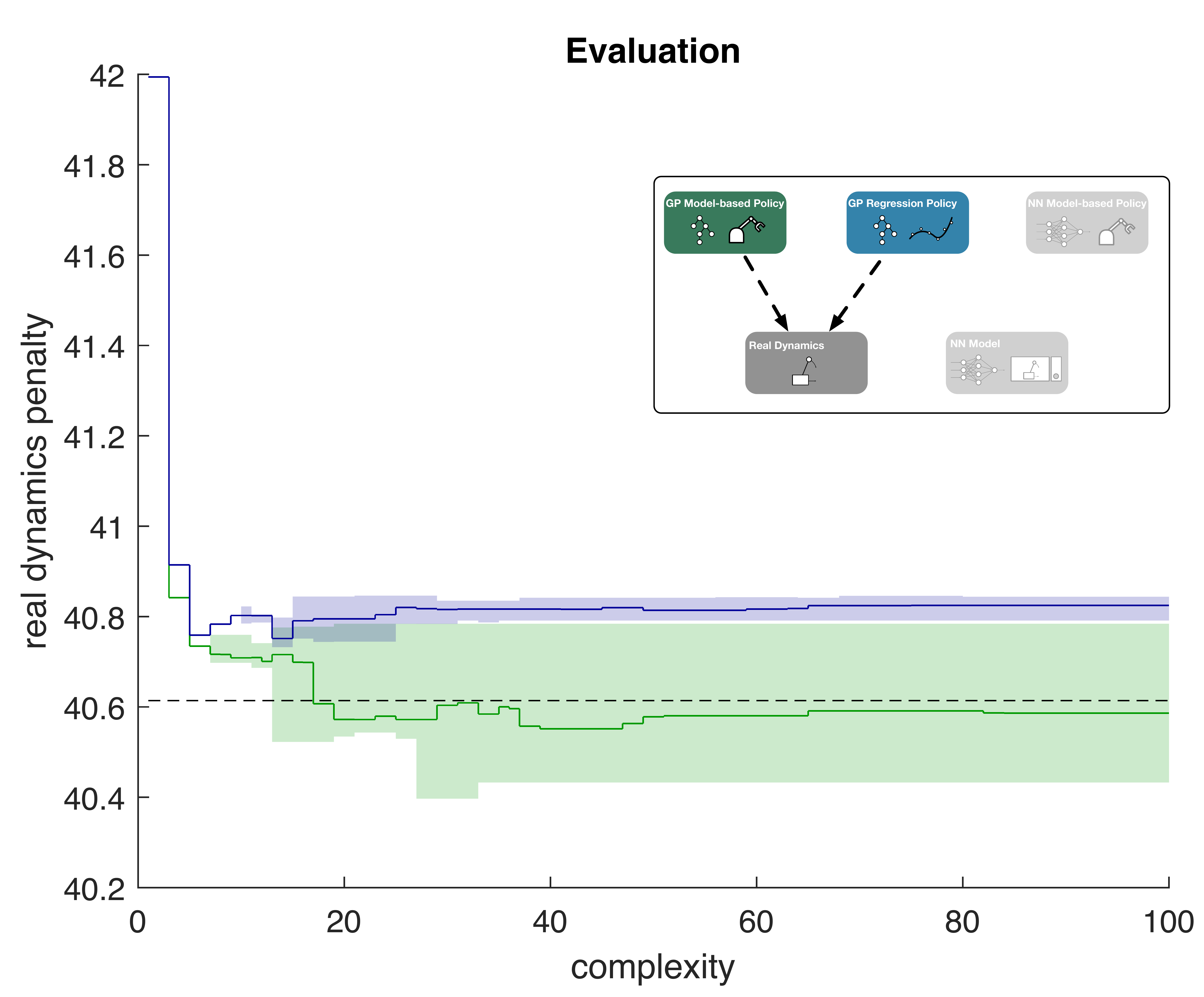}
		\label{mc_pareto_sim}
	}\\
	\subfloat[CPB]{
		\includegraphics[trim=0 0 0 90,clip,width=3in]{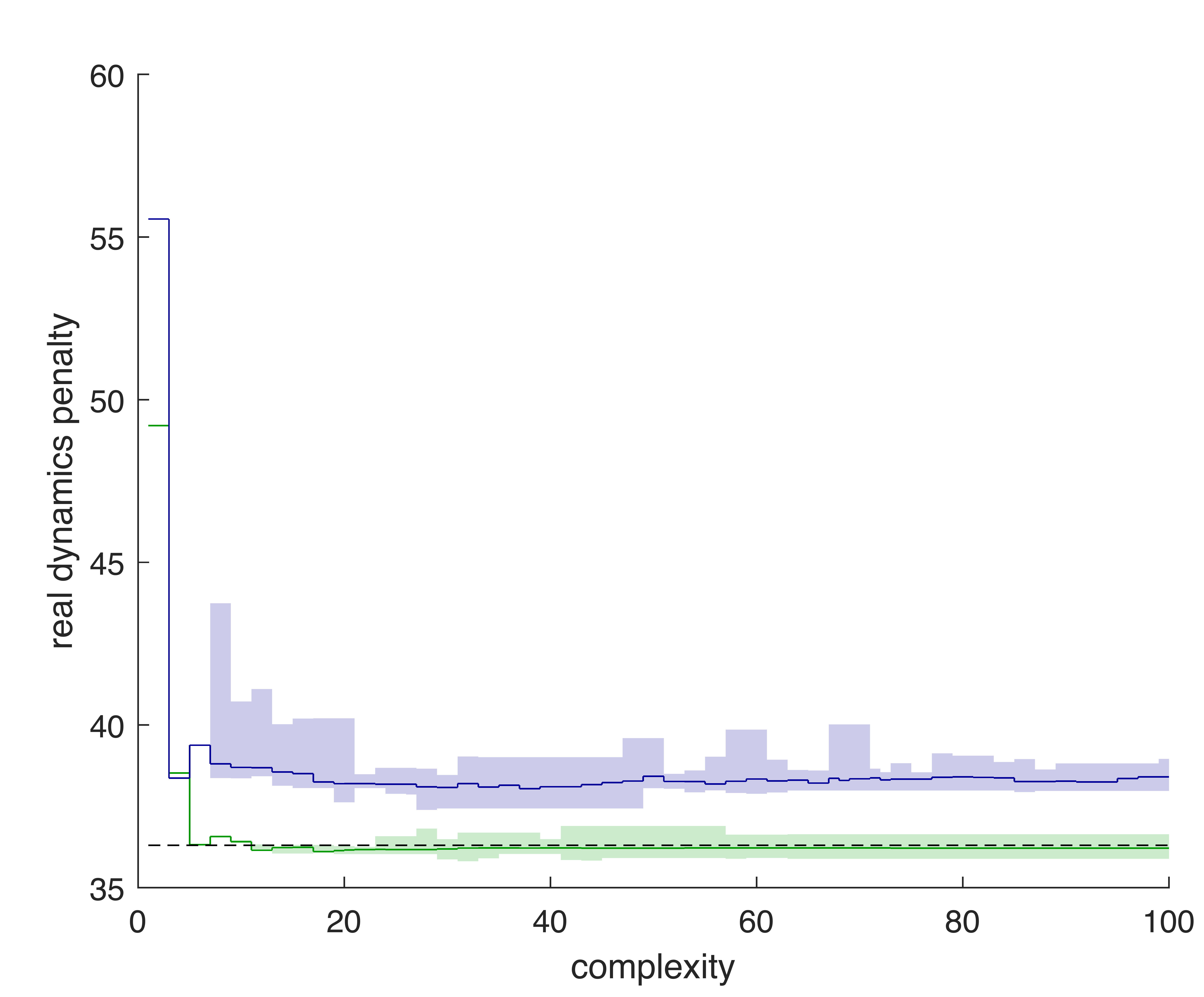}
		\label{cpb_pareto_sim}
	}\\
	\subfloat[IB]{
		\includegraphics[trim=0 0 0 50,clip,width=3in]{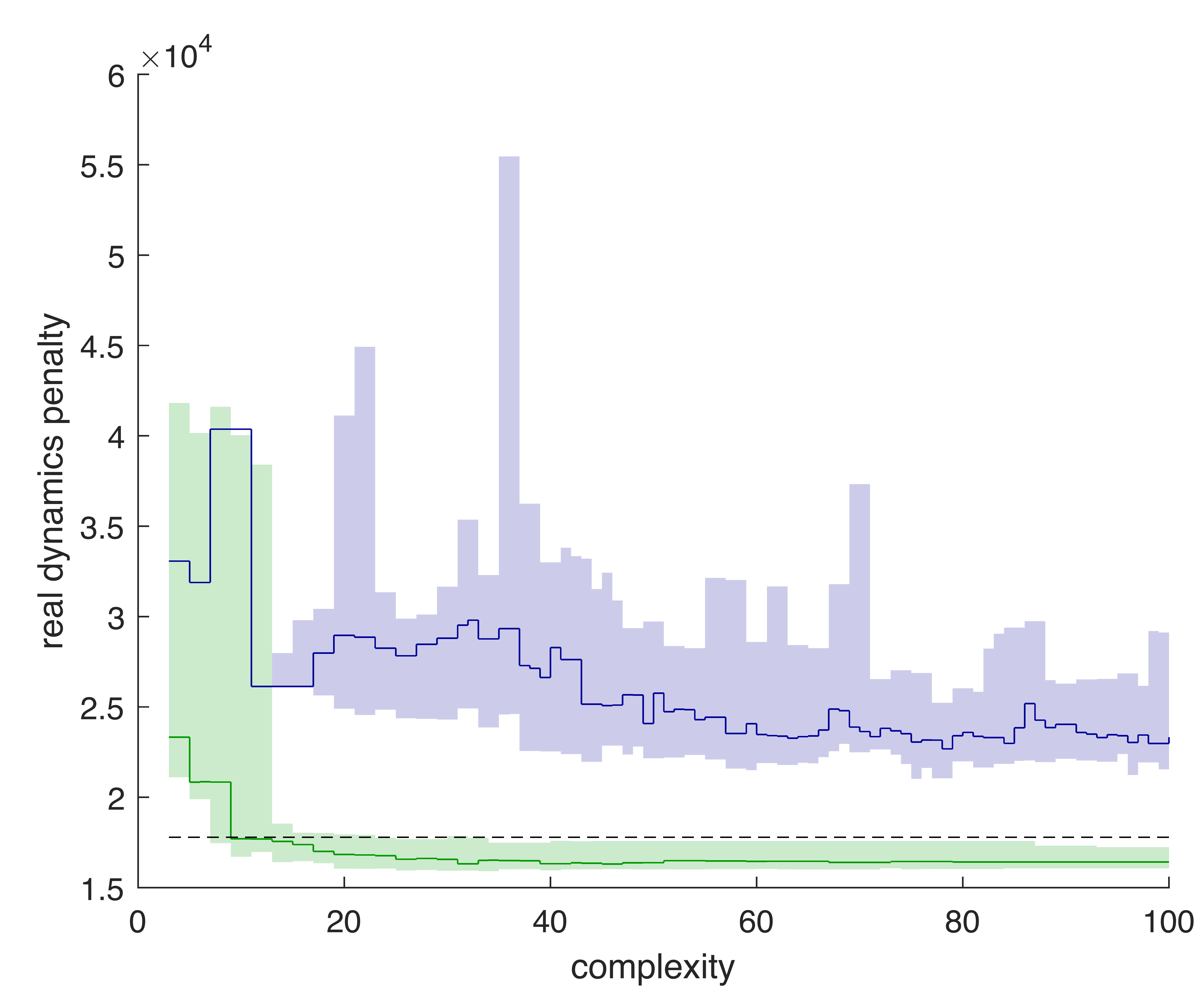}
		\label{ib_pareto_sim}
	}
	\caption{Squashed Pareto fronts from evaluating both GP results with a different set of states on the real benchmark dynamics.
		On average, the GPRL individuals (green) outperform the GP regression individuals (blue).}
	\label{pareto_sim}
\end{figure}

In Figure~\ref{mc_pareto_sim} the performances of GPRL and the GP regression approaches are evaluated by testing the policies of their Pareto fronts with different start states on the real MC dynamics.
Note that on average our GPRL approach produced the best interpretable policies for all complexities.
However, the performance of the symbolic regression approach is quite similar, which suggests that for the MC benchmark such a procedure of creating interpretable RL policies is not impossible.

\subsection{Cart-pole Balancing}

We conducted the CPB experiments using a time horizon $T$ of 100 and a discount vector $\gamma$ of 0.97 ($q=0.05$).
For the CPB experiments, a non-interpretable NN policy with fitness value $\mathcal{F}=-27.1$ (equivalent to a penalty value of 27.1) has been trained prior to the GP experiments.
Based on our experience, this performance value represents a successful CPB policy.
The NN policy has two hidden layers with $\tanh$ activation function and 10 hidden neurons on each layer, i.e., complexity 2471.

In Figure~\ref{cpb_pareto_model} the results of the GP model-based training are compared to the NN policy baseline.
Note that all ten independent GP runs produced Pareto fronts of very similar performance.
In comparison to the NN policy, individuals with complexity $<5$ performed significantly worse with respect to the model penalty.
Individuals with complexity $\geq13$ on the other hand yielded a median penalty of 27.5 or below, which corresponds to an excellent CPB policy suitability.

\begin{figure*}
	\centering
	\includegraphics[width=7in]{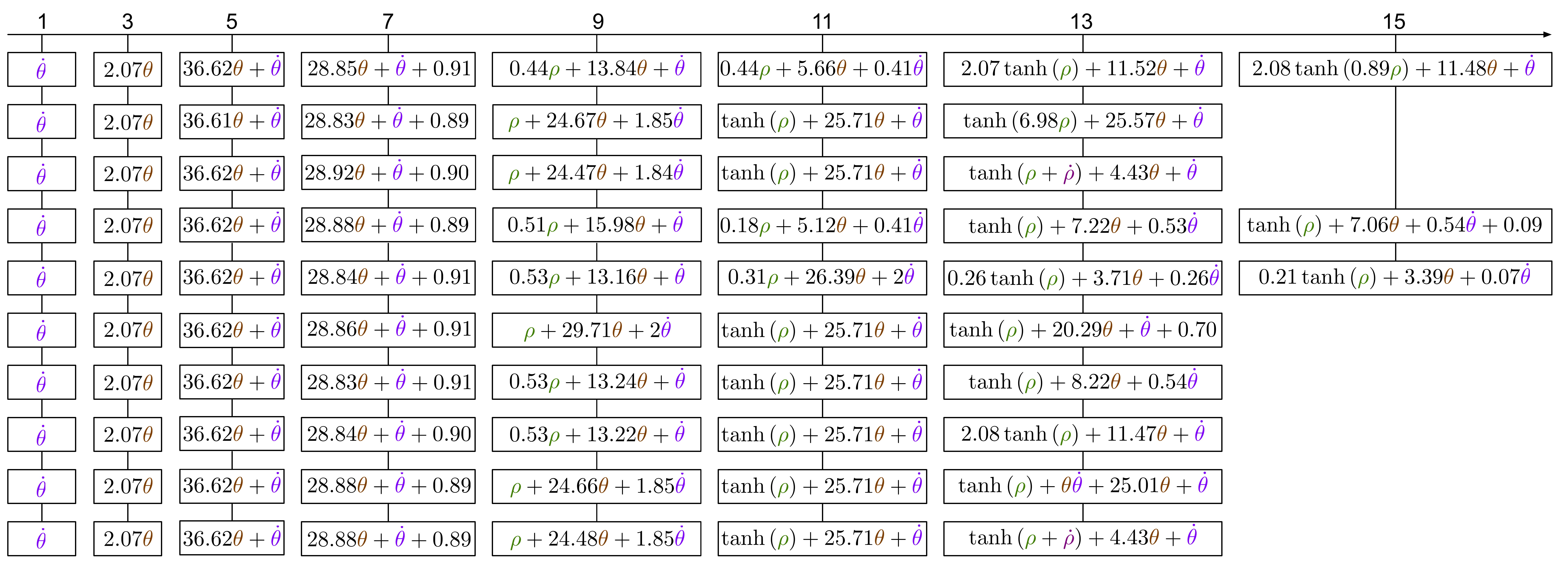}
	\caption{Interpretable CPB policy results by our GPRL approach for complexities 1-15}
	\label{cpb_solutions}
\end{figure*}

Figure~\ref{cpb_solutions} depicts all individuals of the Pareto fronts of the ten experiment runs from complexity 1 to 15.
Note how the solutions agree not only on the utilized state variables but also on the float values of the respective factors.
Differences often only arise due to the multiplication of the whole terms with different factors, i.e., the ratios between the important state variables remain very similar.
Provided with such a policy Pareto chart, experts are more likely to succeed selecting interpretable policies, since common policy concepts are conveniently identified with respect to both, their complexity as well as their model-based performance.

The Pareto front results of the GP regression experiments are presented in Figure~\ref{cpb_pareto_fit}.
Here, the fitness value driving the GP optimization was the regression error with respect to the NN policy.
As expected, the individuals of higher complexity achieve lower errors.
Note that compared to GP model-based training the Pareto fronts results of the 10 experiments are spread throughout a bigger area.
This fact suggests that the NN policy might be highly non-linear in its outputs, which makes it harder for the GP to find solutions in this huge search space.

To evaluate the true performance of the two approaches GP model-based training and GP regression training, the individuals of both sets of Pareto fronts have been tested on the true CPB dynamics.
Figure~\ref{cpb_pareto_sim} shows the resulting \textit{squashed} Pareto fronts compared to the performance of the NN policy.
It is obvious that almost for every complexity the GP model-based approach GPRL is superior to the GP regression idea.
Not only are the median results of significantly lower penalty, but the variance of the GPRL solution is also much lower compared to the GP regression result.
Interestingly, the median GP results for complexity 11 and above even outperformed the NN policy result.
This indicates that the NN policy already overfitted the NN model and exploited its inaccuracies, while the simple GP policy equations generalize better because of their rather restricted structure.

\subsection{Industrial Benchmark}

We conducted the IB experiments using a time horizon $T$ of 100 and a discount vector $\gamma$ of 1 ($q=1$).
For the IB experiments, a non-interpretable NN policy with fitness value $\mathcal{F}=-165.5$ (equivalent to a penalty value of 165.5) has been trained prior to the GP experiments.
Based on our experiences, this performance value represents a successful IB policy.
The NN policy consists of three separate NNs with one hidden layer, $\tanh$ activation functions, and 20 hidden neurons on each layer, i.e., complexity value $43,617$.

The results of ten individual GPRL runs are depicted in Figure~\ref{ib_pareto_model}.
Despite the median model penalty of the Pareto fronts is worse compared to the non-interpretable NN policy, one of the GPRL results produced a slightly better performing policy.
This comes as a surprise, since generally the NN policy has an advantage in degrees of freedom to find the optimal policy.

\begin{figure*}
	\centering
	\includegraphics[width=7in]{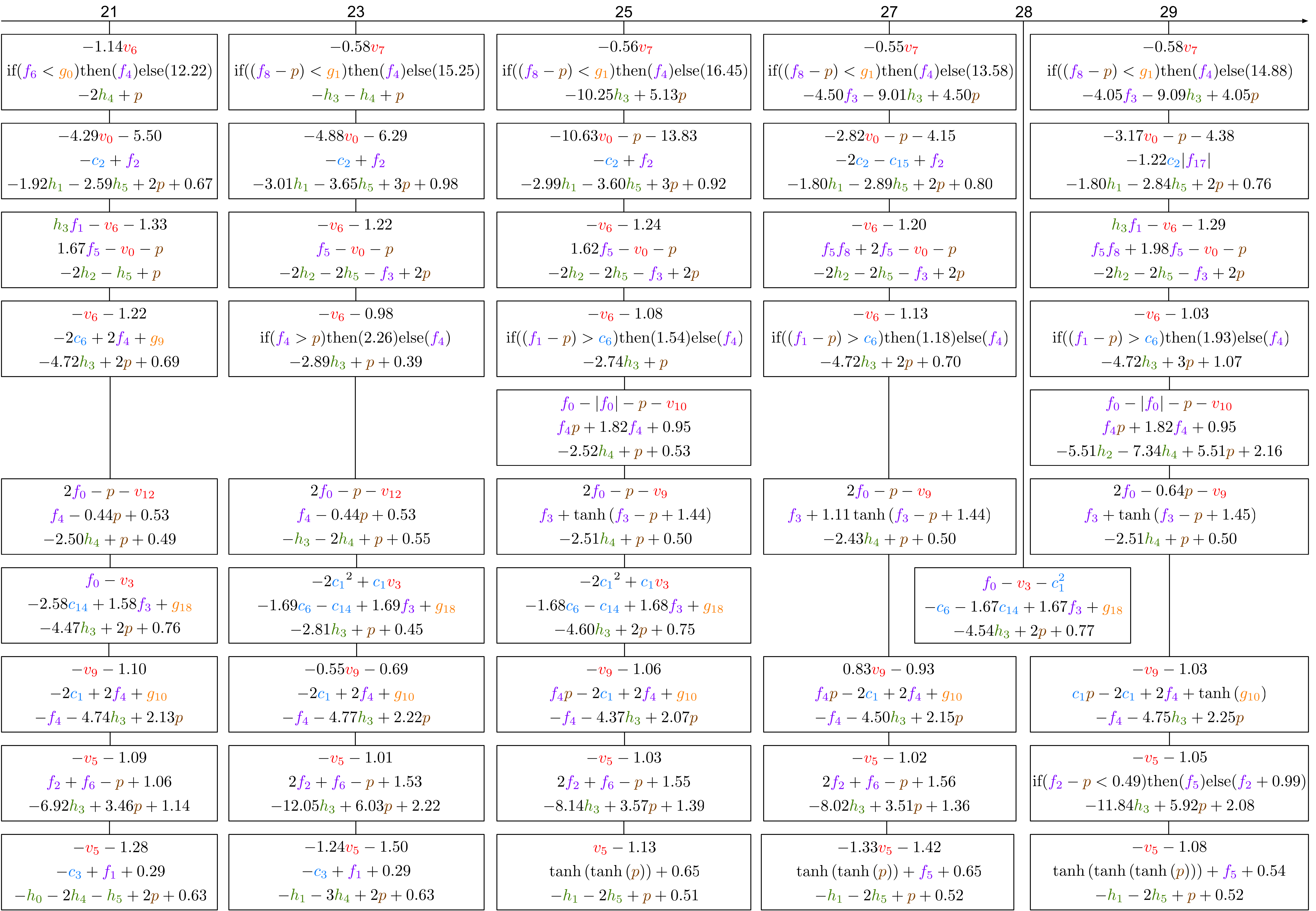}
	\caption{Interpretable IB policies for complexities 21-29.
	The presented 3-dimensional policies are the result of ten independent GPRL trainings.
	Each box contains three policy equations for $\Delta v_0$,$\Delta g_0$, and $\Delta h_0$ (from top to bottom) to calculate actions for time step $t=0$.
	The input variables' indices represent the respective negative time lag in which they have been recorded, e.g., $h_3$ represents the value of shift three time steps ago at $t=-3$.
	The actions are limited to -1 and +1 before they are applied on the system dynamics.
	The input variables $(p,v,g,h,f,c)$ are normalized by subtracting their respective mean $(55.0, 48.75, 50.53, 49.45, 37.51, 166.33)$ and dividing by their respective standard deviation $(28.72, 12.31, 29.91, 29.22, 31.17, 139.44)$.
	These values can easily be calculated from the training data set.}
	\label{ib_solutions}
\end{figure*}

The resulting GPRL policies between complexity 21 and 29 are presented in Figure~\ref{ib_solutions}.
Note that independently learned policies share similar concepts of how policy actions are computed from related state variables with similar time lags.
For example, the equations for $\Delta h$ always use a linear combination of shift value $h$ from time lags -2, -3, or -4 and the constant setpoint value $p$.
Another example is the computation of $\Delta v$, for which a velocity value $v$ with time lags $\geq-10$ is always used.
Moreover, it is possible to reveal common concepts and relevant differences between a rich set of possible solutions.
This presentation could provide the domain experts with important insights on how well-performing policies for the system at hand look like, on which state variables they react, and how they generalize in state space areas where currently insufficient training data is available.

The results of applying GP symbolic regression on a non-interpretable NN policy are shown in Figure~\ref{ib_pareto_fit}.
For each policy action, an independent GP run has been conducted.
After the training multi-dimensional policies have been created in such a way that the accumulated regression errors of $\Delta v$, $\Delta g$, and $\Delta s$ are as low as possible for every possible complexity value.
This procedure has been repeated ten times to yield ten independent IB policies.

In the final step of the experiment, the GPRL and the GP regression solutions have been evaluated on the real IB dynamics.
Figure~\ref{ib_pareto_sim} clearly reveals the strengths of our GPRL approach.
First, the model-based GP training performs significantly better than the symbolic regression training.
Despite even in the MC and CPB experiments GPRL outperformed the regression approach, the experiments with IB  illustrate the complete performance breakdown of the latter.
Second, the good generalization properties of GPRL led to interpretable policies which even outperformed a non-interpretable NN policy from complexity 11 on.
Note that given this result and the superior performance during model-based NN policy training, it can be concluded that the NN policy started to overfit the policy with respect to the NN model penalty.

\section{Conclusion}
\label{chapter:Conclusion}

Our GPRL approach conducts model-based batch RL to learn interpretable policies for control problems on the basis of already existing default system trajectories.
The policies can be represented by compact algebraic equations or Boolean logic terms.
Autonomous learning of such interpretable policies is of high interest for industry domain experts.
Presented with a number of GPRL results for a preferred range of complexity, new concepts for controlling an industrial plant can be revealed.
Moreover, safety concerns can more easily be addressed, if the policy at hand itself, as well as its generalization to certain state space areas, are completely understandable.

The complete GPRL procedure of (i) training a model from existing system trajectories, (ii) learning interpretable policies by GP, (iii) selecting a favorable solution candidate from a Pareto front result has been evaluated for three RL benchmarks, i.e., MC, CPB, and IB.
First, the control performance was compared to a non-interpretable NN policy.
This comparison showed that the GPRL performance on the approximation model can be slightly worse compared to the NN policy result. 
However, when evaluated on the real system dynamics, even interpretable policies of rather low complexity could outperform the non-interpretable approach in many occasions.
This suggests that simple algebraic equations used as policies generalize better on new system states.
In a second evaluation, our GPRL approach has been compared to a straightforward GP utilization as a symbolic regression tool, i.e., fitting the existing non-interpretable NN policy by GP to yield interpretable policies of similar control performance.
All of our experiments showed that this strategy is significantly less suitable to produce policies of adequate performance.

Especially the experiments with the IB indicated that the application of the proposed GPRL approach in industry settings could prove to be of significant interest.
In many cases, data from systems is readily available and interpretable simple algebraic policies are favored over black-box RL solutions, such as non-interpretable NN policies.

\section*{Acknowledgment}

The project on which this report is based was supported with funds from the German Federal Ministry of Education and Research under project number 01IB15001. The sole responsibility for the report's contents lies with the authors.

\section*{References}

\bibliographystyle{elsarticle-harv}
\biboptions{longnamesfirst,authoryear,round,semicolon}
\bibliography{bibliography.bib}


\end{document}